\documentclass[11pt]{article}

\usepackage[preprint]{acl}

\usepackage{times}
\usepackage{latexsym}
\usepackage{enumitem}
\usepackage[T1]{fontenc}

\usepackage[utf8]{inputenc}

\usepackage{microtype}
\usepackage{algorithm}
\usepackage{algorithmic}
\usepackage{multirow}
\usepackage{inconsolata}
\usepackage{marvosym}
\newcommand{\coloremojicode}[1]{\Letter}
\usepackage{graphicx}

\usepackage{amsmath}
\usepackage{amssymb}
\usepackage{tcolorbox}
\usepackage{array}
\usepackage{tabularx}
\usepackage{booktabs}
\usepackage{subcaption}
\usepackage{graphicx}
\usepackage{courier}
\usepackage{xcolor} 

\definecolor{darkgreen}{RGB}{0, 150, 0} 

\newcommand{\pos}[1]{\scriptsize{\textcolor{darkgreen}{(+#1)}}}

\title{Learning the Boundary of Solvability:\\Aligning Large Language Models to Detect Unsolvable Problems}

\author{
 Dengyun Peng$^{1,3}$\thanks{Equal contribution} \quad 
 Qiguang Chen$^{1}$\footnotemark[1] \quad
 Bofei Liu$^{1}$ \quad
 Jiannan Guan$^{1}$ \quad 
 Libo Qin$^{2}$\thanks{Corresponding Author} \\
 \textbf{Zheng Yan}$^{1}$ \quad
\textbf{Jinhao Liu}$^{1}$ \quad
\textbf{Jianshu Zhang}$^{3}$ \quad 
 \textbf{Wanxiang Che$^{1}$}\footnotemark[2] \\[2mm]
 $^{1}$LARG, Research Center for Social Computing and Interactive Robotics, HIT \\
 $^{2}$School of Computer Science and Engineering, Central South University \\
 $^{3}$iFLYTEK 
}

\begin{document}
\maketitle
\begin{abstract}
Ensuring large language model (LLM) reliability requires distinguishing \textit{objective unsolvability} (inherent contradictions) from \textit{subjective capability limitations} (tasks exceeding model competence). Current LLMs often conflate these dimensions, leading to hallucinations in which they return confident answers to inherently unsolvable queries. To address this issue, we propose a multi-domain dataset containing both solvable and unsolvable questions, \textbf{UnsolvableQA}, together with an alignment framework, \textbf{UnsolvableRL}. First, we construct UnsolvableQA by ``Reverse Construction'' that systematically injects logical contradictions into otherwise valid reasoning chains. Second, we introduce UnsolvableRL, a reinforcement learning paradigm that balances objective unsolvability detection with calibrated confidence under capability limits. Empirically, our approach achieves robust unsolvability detection ($>\!85\%$ detection rate) and boosts solvable reasoning accuracy from 43.4\% to 69.4\% on Qwen3-4B-Instruct. Crucially, we identify a data–training interaction: strict alignment constraints induce \textit{Capability Collapse} without unsolvable data, but act as a regularizer for rigor when such data are included, thereby improving overall robustness. Our code and data are available at \url{https://github.com/sfasfaffa/unsolvableQA}.
\end{abstract} 

\section{Introduction}

As Large Language Models (LLMs) achieve strong performance on complex reasoning tasks~\citep{chen2025towards,wei2022chain,guo2025deepseek,jaech2024openai,ICLR2024_d53538ba}, a key research direction \citep{qin2025largelanguagemodelsmeet} is ensuring they recognize their own limitations and behave reliably. Earlier work on confidence calibration~\citep{xiong2023can,li2024confidence,yoon2025reasoning,fisch2022calibrated} during reasoning underscores the need for models to know when not to answer. Further, studies on LLMs' \textbf{capability boundaries}~\citep{NEURIPS2024_62ab1c2c} examine the self-awareness of their limitations to avoid unproductive reasoning on problems they are likely to fail~\citep{zhang2025self,mei2025reasoning,kalai2025languagemodelshallucinate,chen2025aware}. More recently, TruthRL~\citep{wei2025truthrl} has encouraged LLMs to refuse when uncertain through a ternary reward mechanism.

\begin{figure}[t]
    \centering
    \includegraphics[width=\linewidth]{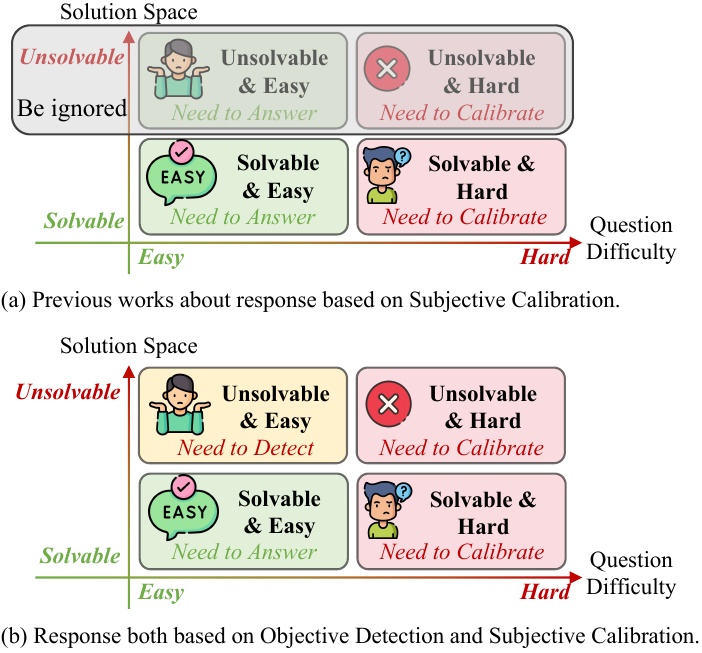}
    \caption{Comparison of alignment paradigms. We treat the solution space (solvable vs. unsolvable) and question difficulty (easy vs. hard) as orthogonal dimensions, requiring models to \textbf{objectively detect} solvability and \textbf{subjectively calibrate} confidence.}
    \label{fig:intro_quadrant}
\end{figure}

However, a key limitation of these approaches is that they mainly focus on refusing questions that are \textit{solvable but currently beyond the model's capabilities} (the ``Need to Calibrate'' region in Figure~\ref{fig:intro_quadrant} (a)). This emphasis overlooks a complementary, and more basic, challenge: identifying problems that are \textit{inherently unsolvable because they contain logical contradictions}. More problematically, existing LLMs often misinterpret such unsolvability as high difficulty and may hallucinate answers even when the contradiction is explicit.

To address this, our work treats \textit{Solution Space} (solvable vs. unsolvable) and \textit{Question Difficulty} (easy vs. hard) as orthogonal dimensions (Figure~\ref{fig:intro_quadrant} (b)). Concretely, we argue that a reliable AI system must exhibit two behaviors: (1) \textbf{Objective Detection}, identifying internally contradictory problems and flagging them as ``unsolvable'' (yellow area); and (2) \textbf{Subjective Calibration}, safely refusing theoretically solvable but ``difficult'' problems that exceed the model’s reasoning capability. Without this distinction, models may hallucinate, forcing answers for inherently unsolvable problems.

First, to support rigorous analysis of unsolvability, we develop a fully automated synthesis pipeline and introduce \textbf{UnsolvableQA}, a dataset of paired solvable and unsolvable instances spanning puzzles and mathematics. For mathematics, inspired by \citet{lightman2024lets,ling2023deductive}, we propose a \textit{``Reverse Construction''} method that injects controlled contradictions into valid rationales to generate matching unsolvable problems. For programmatically verifiable puzzles, we construct instances via code-level logical contradictions. Notably, our logic puzzle pipeline enables the rapid generation of infinite instances with arbitrary difficulty, and we will fully release our construction code upon acceptance.

Then, to train models that remain robust on both solvable and unsolvable inputs, we propose \textbf{UnsolvableRL}, a reinforcement-learning (RL) method based on GRPO~\citep{shao2024deepseekmath}. Unlike prior approaches that conflate refusal types, we define 3 distinct RL objectives: (1) {Inherent unsolvability} rewards generating an explicit refusal tag when solvability is detected; (2) {False unsolvability Penalty} discourages erroneous refusals by penalizing labeling solvable problems as unsolvable, thereby requiring explicit solvability checks before refusal; (3) {Capability calibration} uses a mechanism that rewards refusal only when the problem difficulty exceeds the LLMs' current capability.

Empirical results show a synergistic effect: UnsolvableRL elevates the unsolvable problem detection rate of Qwen3-4B from 38.8\% to 87.5\% and boosts its solvable problem accuracy from 43.4\% to 69.4\%, achieving an overall score of 78.6\%. Our ablation study reveals three key insights: (1) models cannot learn to detect unsolvable problems without explicit unsolvable training data; (2) penalizing false refusals without such data causes “capability collapse”, where models stop refusing altogether; (3) when paired with unsolvable data, the penalty acts as a regularizer, trading a small drop in detection for stronger reasoning accuracy, a Pareto improvement over baselines.

Our contributions are summarized as follows:
\begin{itemize}[leftmargin=16pt, itemsep=0pt, topsep=0pt]
    \item We first propose that reasoning alignment should treat objective unsolvability and subjective capability as orthogonal axes for reliable reasoning with clear decision boundaries.
    \item We introduce \textbf{UnsolvableQA}, a dataset built with a ``Reverse Construction'' pipeline that injects logical contradictions into valid rationales to create verifiable unsolvable instances.
    \item We propose \textbf{UnsolvableRL}, an RL paradigm with dynamic confidence thresholds and penalties for incorrect refusals, enabling models to solve feasible tasks, detect solvability, and calibrate confidence on hard inputs.
    \item Controlled ablation studies validate the necessity and synergy of our two key designs: training on unsolvable data and applying a calibrated refusal penalty. This combination drives a Pareto improvement, maintaining high accuracy on solvable tasks while robustly detecting unsolvable ones.
\end{itemize}

\section{Related Work}
\label{sec:related_work}

There are two primary concerns related to our work: automated dataset construction for reasoning and reinforcement learning approaches for incentivizing selective refusal.\vspace{-5pt}

\paragraph{Automatic Reasoning Dataset Construction.} As the reliance of Large Language Models on data quality increases, Automated Dataset Construction (ADC) has emerged as a prominent research direction~\citep{liu2024automatic, wang2023data}.
In the realm of reasoning, recent efforts on automatic data construction primarily focus on ensuring data solvability and verifiability through formal or programmatic means.
Representative works such as Enigmata~\citep{chen2025enigmata} and BIGGSM~\citep{NEURIPS2024_62ab1c2c} have significantly enriched the resources for solvable reasoning tasks by employing synthetic puzzles and difficulty-graded annotations, respectively.
Concurrently, research on verifying the correctness of rationale traces has also been advancing~\citep{ling2023deductive, chen2025towards}.

In contrast, progress in dataset construction regarding problem unsolvability has been relatively slow.
Earlier, SQuAD 2.0~\citep{rajpurkar2018knowdontknowunanswerable} examined whether questions are answerable given a long context, but it does not adequately evaluate refusal behavior in complex logical-reasoning settings.
Consequently, for reasoning tasks, existing data-synthesis efforts focus mainly on \textit{solvable} instances while largely neglecting \textit{unsolvable} questions, leaving a significant research gap.\vspace{-5pt}

\begin{figure*}[t]
    \centering
    \includegraphics[width=\linewidth]{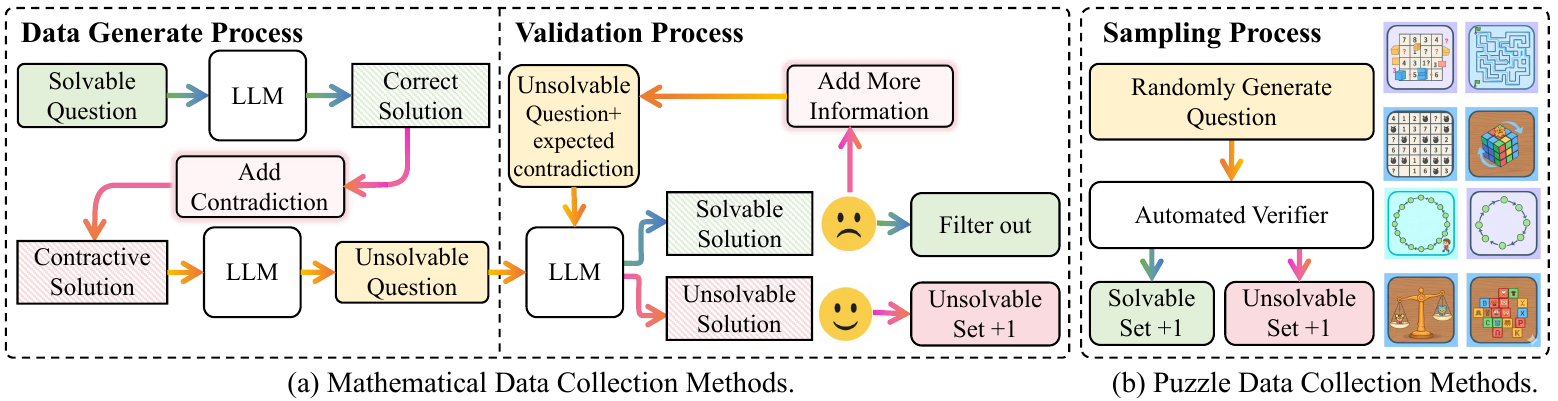}
    \caption{The \textbf{UnsolvableQA} data collection pipeline. \textbf{(a) Mathematical Data Collection:} It employs a ``Reverse Construction'' method where contradictions are injected into the solution path of solvable problems. The resulting instances are verified by an LLM to ensure they lead to a contradiction (unsolvable) rather than a valid solution. \textbf{(b) Puzzle Data Collection:} It utilizes domain-specific programmatic generators and automated deterministic verifiers (e.g., SAT solvers, DFS) to rigorously classify randomly generated instances into solvable and unsolvable sets.}
    \label{fig:pipeline}
\end{figure*}

\paragraph{Reinforcement Learning for Reasoning.} Beyond dataset construction, Reinforcement Learning (RL)~\citep{ouyang2022traininglanguagemodelsfollow,schulman2017proximalpolicyoptimizationalgorithms,rafailov2024directpreferenceoptimizationlanguage} has become a central approach for aligning LLMs with desired behaviors such as safety and truthfulness.
Recent RL algorithms for reasoning, such as GRPO~\citep{shao2024deepseekmath} and DAPO~\citep{yu2025dapoopensourcellmreinforcement}, further enhance models’ reasoning abilities~\citep{sheng2025hybridflow}.
Concurrent work also leverages RL to improve reasoning efficiency under cost constraints~\citep{tu2025learning, team2025kimi, guo2025deepseek, wang2025aspoasymmetricimportancesampling, chen2025aware}.
In parallel, advances such as TruthRL~\citep{wei2025truthrl} introduce objectives that encourage models to refuse likely hallucinations, and research on self-awareness enables models to decline answering when appropriate~\citep{zhang2025self, mei2025reasoning,kalai2025languagemodelshallucinate,Steyvers2025}.

Despite these advancements, existing frameworks largely conflate \textit{objective unsolvability} with \textit{subjective uncertainty}. By focusing primarily on solvable instances or confidence calibration, they overlook the necessity of explicitly learning to identify logical contradictions. We bridge this gap by systematically constructing guaranteed-unsolvable data to enforce a robust three-way decision boundary: solving feasible tasks, detecting inherent unsolvability, and calibrating refusal for problems beyond the model's current capabilities.
\section{UnsolvableQA Construction}
\label{sec:dataset}

Existing benchmarks mainly target solvable tasks, so models remain prone to hallucinations when assessing feasibility. To address this, we introduce UnsolvableQA, a dataset for detecting intrinsic contradictions in problem statements, using a dual-track methodology (in Figure~\ref{fig:pipeline}) aligned with domain-specific verification constraints.

\subsection{Mathematics Problem Generation via Reverse Construction}
\label{sec:math_gen}

For mathematical reasoning tasks (e.g., algebra, geometry, combinatorics from AIME datasets), direct programmatic verification is often infeasible. We introduce a \textbf{Reverse Construction} methodology (see Figure~\ref{fig:pipeline} (a)) that leverages large language models to generate unsolvable problems by injecting logical contradictions into solvable seed problems, yielding high-quality unsolvable mathematics problems for training unsolvable policies.

Our reverse construction pipeline operates through two sequential stages:

\vspace{0.3em}
\noindent\textbf{Data Generation Process.}
First, we validate each seed problem by confirming that a strong reasoning model (DeepSeek-Reasoner) can correctly solve it and produce a complete chain-of-thought, ensuring sufficient reasoning capability. Then, we employ DeepSeek-Reasoner to design logical contradictions via two complementary strategies: \textit{Constraint Contradiction}, which introduces conflicting external conditions, and \textit{Axiom Contradiction}, which modifies reasoning steps to violate fundamental mathematical axioms. The model generates a structured plan specifying the location and mechanism of contradiction injection.

\vspace{0.3em}
\noindent\textbf{Validation Process.}
Based on this plan, LLMs then reverse-engineers a new problem statement that preserves the stylistic and complexity characteristics of the seed problem while embedding the designed contradiction. Finally, we implement a two-tier verification protocol: the generated problem is first presented to DeepSeek-R1 to autonomously detect contradictions; if model fails to independently identify the contradiction, we proceed to the second tier. In this stage, we explicitly provide the model with both the final unsolvable problem statement and the predefined contradiction plan, thereby supplementing it with the intended contradiction information beyond the problem text alone. The model is then prompted to reassess the problem’s solvability given this complete contextual guidance. Only instances verified as unsolvable are retained, and we manually verified the final test set.

\subsection{Logic Puzzle Generation}
\label{sec:logic_gen}

To acquire strictly unsolvable questions in logic puzzle scenarios, we implement automated generators for four representative logic domains, following the sampling and verification process in Figure~\ref{fig:pipeline} (b). Each domain uses a deterministic solver to verify solutions and confirm unsolvability.\vspace{-5pt}

\paragraph{Game24.} 
Solvability is checked by exhaustively enumerating all binary expression trees with exact rational arithmetic (Python \texttt{Fraction}) to avoid floating-point error. Unsolvable instances are obtained by \textit{rejection sampling}, retaining number sets for which the exhaustive search proves that no solution exists.\vspace{-5pt}

\paragraph{Hamiltonian Cycle/Path.} 
We encode graph traversability as a Boolean Satisfiability problem and verify randomly generated graphs using SAT solvers (MiniSat backend via \texttt{pysat}). Unsolvable instances are constructed by injecting structural impossibilities, such as removing critical edges to create bottlenecks or disconnecting components.\vspace{-5pt}

\paragraph{Hitori.} 
We formulate Hitori as a constraint satisfaction problem and, using \texttt{python-constraint} to generate puzzles with unique solutions. For unsolvable instances, we inject conflicting constraints, for example forcing a row or column to contain duplicate numbers that cannot be removed without violating adjacency rules.\vspace{-5pt}

\paragraph{Maze Navigation.} 
Base mazes are generated via randomized Depth-First Search (DFS) to ensure nontrivial structure. To obtain unsolvable instances, we analyze all valid paths in a solvable maze and place an obstacle at a \textit{critical junction} shared by every path, then verify unsolvability using Breadth-First Search (BFS).

\begin{table}[t]
    \centering

    \resizebox{\linewidth}{!}{%
    \begin{tabular}{llccc}
        \toprule
        \textbf{Split} & \textbf{Domain} & \textbf{Solvable} & \textbf{Unsolvable} & \textbf{Total} \\
        \midrule
        Train & Game24 & 50 & 50 & 100 \\
        Train & Hamiltonian Cycle & 48 & 48 & 96 \\
        Train & Hamiltonian Path & 50 & 50 & 100 \\
        Train & Hitori & 50 & 50 & 100 \\
        Train & Maze & 100 & 59 & 159 \\
        Train & AIME (1983--2023) & 50 & 32 & 82 \\
        \midrule
        Test & Game24 & 50 & 50 & 100 \\
        Test & Hamiltonian Cycle & 48 & 50 & 98 \\
        Test & Hamiltonian Path & 50 & 50 & 100 \\
        Test & Hitori & 50 & 50 & 100 \\
        Test & Maze(Easy) & 100 & 94 & 194 \\
        Test & Maze(Hard) & 100 & 100 & 200 \\
        Test & AIME (2024--2025) & 60 & 47 & 107 \\
        \midrule
        Train Total & -- & 348 & 289 & 637 \\
        Test Total & -- & 458 & 441 & 899 \\
        \bottomrule
    \end{tabular}}
    \caption{Unified vertical statistics for \textbf{UnsolvableQA} train/test partitions. Train Maze has no difficulty split; AIME year ranges differ.}
    \vspace{-5pt}
    \label{tab:dataset_stats}
\end{table}
\subsection{Dataset Statistics and Quality Control}
We construct \textbf{UnsolvableQA} to maintain approximately balanced instances per domain (Table~\ref{tab:dataset_stats}). The dataset comprises 899 test problems (458 solvable, 441 unsolvable) and 637 training problems (348 solvable, 289 unsolvable). While training prompts follow templated formats and pool Maze difficulties, test prompts are diversified and preserve fine-grained splits (including distinct AIME years\footnote{AIME year ranges: train 1983--2023; test 2024--2025.}). All instances undergo rigorous quality control: logic puzzles are verified via SAT/CSP/graph solvers, mazes via deterministic search, and math problems through multi-pass LLM verification and manual checks. We also introduce two OOD testsets in Appendix~\ref{sec:appendix_ood}.

\section{UnsolvableRL Methodology}

\begin{figure*}[t]
    \centering
    \includegraphics[width=\linewidth]{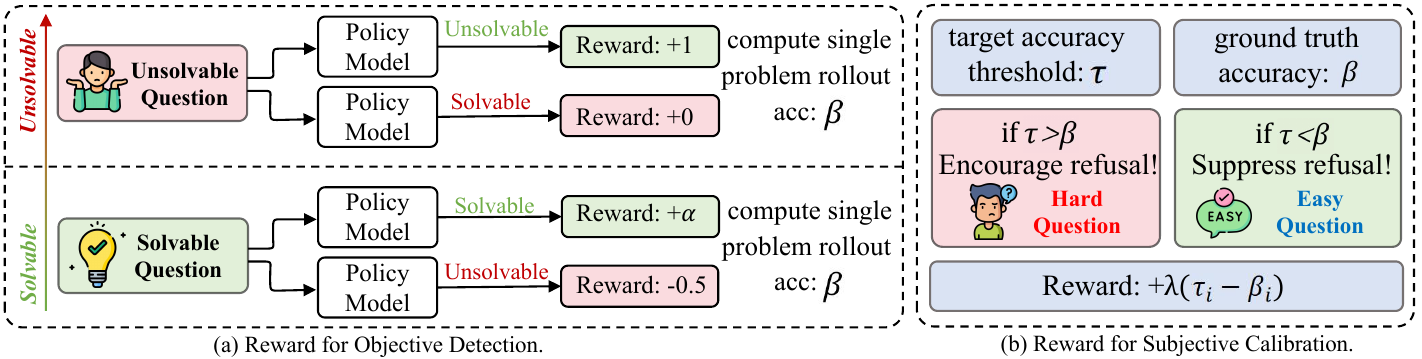}
    \caption{UnsolvableRL framework with two orthogonal components: \textit{(a) Unsolvability Detection}, which penalizes labeling solvable queries as unsolvable to preserve reasoning incentives.
    \textit{(b) Capability Calibration}, which regulates refusal by comparing the target threshold (\(\tau\)) with rollout accuracy (\(\beta\)); it encourages refusal when \(\tau > \beta\) and discourages it when \(\tau < \beta\) via \(\lambda(\tau_i - \beta_i)\).}
    \label{fig:reward_design}
\end{figure*}

\subsection{Problem Formulation}
We formulate the reasoning task as generating a response \(y\) for a problem instance \(x\), sampled from a distribution \(\mathcal{D}\). Each instance $x\in \mathcal{D}$ is either objectively solvable (\(x \in \mathcal{S}\)) or unsolvable (\(x \in \mathcal{U}\)), and is further categorized as subjectively easy (\(x \in \mathcal{E}\)) or hard (\(x \in \mathcal{H}\)).

Specifically, to satisfy three distinct behavioral objectives, our training framework optimizes a policy \(\pi_\theta\) as follows. First, for solvable and subjectively easy instances within the model’s capabilities (\(x \in \mathcal{S} \cap \mathcal{E}\)), the objective is \textbf{Reasoning Accuracy}, requiring the model to produce correct reasoning chains and final answers. Second, for inherently unsolvable instances (\(x \in \mathcal{U} \cap \mathcal{E}\)), the objective is \textbf{Unsolvability Detection}: the model must explicitly recognize the inconsistency and output a dedicated tag (e.g., \texttt{<unsolvable>}) instead of hallucinating a solution, which we treat as an \textit{objective correctness} task. Third, for subjectively hard yet solvable instances (\(x \in \mathcal{H}\)) that exceed the model’s effective reasoning capacity (as indicated by low internal confidence), the objective is \textbf{Capability Calibration}, where the model should prudently abstain (e.g., output \texttt{<beyond\_capacity>}) to reduce false positives.

\subsection{Group-Relative Policy Optimization}
To efficiently optimize the policy across heterogeneous objectives without training a separate value function, this work adopts Group Relative Policy Optimization (GRPO). For each input $x$, the policy samples a group of $G$ outputs $\mathcal{Y} = \{y_1, y_2, \dots, y_G\}$. The optimization objective is defined as:
\begin{equation}
\begin{split}
\mathcal{J}(\theta) &= \mathbb{E}_{x, \mathcal{Y}} \bigg[ \frac{1}{G} \sum_{i=1}^{G} \min \Big( r_i(\theta) A_i, \\
&\qquad \text{clip}(r_i(\theta), 1-\epsilon, 1+\epsilon) A_i \Big) \bigg],
\end{split}
\label{eq:grpo_obj}
\end{equation}
where $r_i(\theta) = \frac{\pi_\theta(y_i|x)}{\pi_{\theta_{\text{old}}}(y_i|x)}$ is the probability ratio, and $\epsilon$ is the clipping parameter. GRPO computes the advantage $A_i$ for each output $y_i$ by normalizing rewards within the group instead of using a critic model:
\begin{equation}
A_i = \frac{R(y_i \mid x) - \mu_{\mathbf{R}}}{\sigma_{\mathbf{R}} + \epsilon},
\label{eq:grpo_adv}
\end{equation}
where $\mu_{\mathbf{R}}$ and $\sigma_{\mathbf{R}}$ are the mean and standard deviation of the rewards $\mathbf{R}=\{R(y_j \mid x)\}_{j=1}^G$ within the group. This group-based reward facilitates efficient and stable training, helps effectively distinguish different responses to the same problem, and eliminates the need for critic and reward models that occupy massive GPU memory.

\subsection{Reward Design}

\begin{table*}[t]
\centering
\small
\resizebox{\textwidth}{!}{
\begin{tabular}{ll c c c c c c c}
\toprule
 & & \multicolumn{3}{c}{\textbf{SOTA Baselines}} & \multicolumn{2}{c}{\textbf{Qwen3-1.7B}} & \multicolumn{2}{c}{\textbf{Qwen3-4B}} \\
\cmidrule(lr){3-5} \cmidrule(lr){6-7} \cmidrule(lr){8-9}
\textbf{Dataset} & \textbf{Metric} & Deepseek-V3.2-R & GPT-5.1-Low & Gemini-3 & Instruct & + UnsolvableRL & Instruct & + UnsolvableRL \\
\midrule
\multirow{2}{*}{Game24} 
 & S / U & \textbf{98.0} / 77.5 & 14.0 / 50.0 & 94.0 / 94.0 & 78.0 / 23.0 & 84.0 / \textbf{100.0} & 81.0 / 17.0 & 94.5 / 99.0 \\
 & Mean & 87.8 & 32.0 & 94.0 & 50.5 & 92.0 \pos{41.5} & 49.0 & \textbf{96.8} \pos{47.8} \\
\midrule
\multirow{2}{*}{HamCycle} 
 & S / U & 83.5 / 94.0 & 8.3 / 82.0 & \textbf{100.0} / \textbf{98.0} & 22.9 / 28.0 & 20.8 / 88.0 & 37.5 / 57.0 & 41.1 / 94.5 \\
 & Mean & 88.8 & 45.1 & \textbf{99.0} & 25.5 & 54.4 \pos{28.9} & 47.3 & 67.8 \pos{20.5} \\
\midrule
\multirow{2}{*}{HamPath} 
 & S / U & 84.0 / 96.0 & 50.0 / 86.0 & \textbf{100.0} / 90.0 & 14.0 / 33.0 & 27.0 / 85.0 & 37.0 / 56.5 & 55.0 / \textbf{96.5} \\
 & Mean & 90.0 & 68.0 & \textbf{95.0} & 23.5 & 56.0 \pos{32.5} & 46.8 & 75.8 \pos{29.0} \\
\midrule
\multirow{2}{*}{Hitori} 
 & S / U & 70.0 / 86.0 & 6.0 / 12.0 & \textbf{98.0} / \textbf{100.0} & 8.0 / 21.0 & 11.0 / 59.0 & 34.5 / 6.5 & 63.5 / 94.5 \\
 & Mean & 78.0 & 9.0 & \textbf{99.0} & 14.5 & 35.0 \pos{20.5} & 20.5 & 79.0 \pos{58.5} \\
\midrule
\multirow{2}{*}{Maze (Easy)} 
 & S / U & \textbf{100.0} / 98.9 & 95.0 / \textbf{100.0} & 99.0 / 94.6 & 0.0 / 85.6 & 0.0 / 95.2 & 32.7 / 55.6 & 96.5 / 98.9 \\
 & Mean & \textbf{99.5} & 97.5 & 96.8 & 42.8 & 47.6 \pos{4.8} & 44.1 & 97.7 \pos{53.6} \\
\midrule
\multirow{2}{*}{Maze (Hard)} 
 & S / U & \textbf{98.0} / \textbf{98.0} & 86.0 / 94.0 & 96.0 / 91.0 & 0.0 / 80.0 & 0.0 / 79.0 & 13.3 / 64.5 & 65.8 / 89.0 \\
 & Mean & \textbf{98.0} & 90.0 & 93.5 & 40.0 & 39.5 \scriptsize{(-0.5)} & 38.9 & 77.9 \pos{39.0} \\
\midrule
\multirow{2}{*}{AIME 24-25} 
 & S / U & 85.0 /40.3 & 61.7 /  \textbf{42.5} & \textbf{95.0} / 21.2 & 38.3 / 21.2 & 35.4 / 28.7 & 67.9 / 14.8 & 69.6 / 40.4 \\
 & Mean & \textbf{62.7} & 52.1& 58.1 & 29.8 & 32.1 \pos{2.3} & 41.4 & 55.0 \pos{13.6} \\
\midrule
\multirow{2}{*}{\textbf{Overall}} 
 & \textbf{S / U} & 88.4 / 84.4 & 45.9 / 66.6& \textbf{97.4} / 84.1 & 23.0 / 41.7 & 25.5 / 76.4 & 43.4 / 38.8 & 69.4 / \textbf{87.5} \\
 & \textbf{Mean} & 86.1 & 56.2 & \textbf{90.8} & 32.4 & 50.9\pos{18.5} & 41.1 & 78.6 \pos{37.5} \\
\bottomrule
\end{tabular}}
\caption{Detailed performance comparison. For each dataset, the first row reports \textbf{Solvable Accuracy (S) / Unsolvable Detection Rate (U)}, and the second row reports their \textbf{Mean (M)}. Values in parentheses indicate the absolute improvement over the base Instruct model. We bold the \textbf{global best score} across all models. Qwen3-4B + UnsolvableRL achieves SOTA performance in Unsolvable Detection (Overall U) and Game24 Mean accuracy.}
\label{tab:vertical_main_results_compact}
\end{table*}
As illustrated in Figure~\ref{fig:reward_design}, our framework explicitly distinguishes between three types of model responses: (1) \textit{\textbf{Reward for Reasoning Accuracy}} (providing a solution), (2) \textit{\textbf{Reward for Unsolvability Detection}} (identifying inherent contradictions), and (3) \textit{\textbf{Reward for Capability Calibration}} (admitting capability limits). We design a composite reward function that applies distinct mechanisms to these behaviors to align objective correctness with subjective capability.\vspace{-5pt}

\paragraph{Reward for Reasoning Accuracy ($R_{\text{acc}}$).} 
It evaluates the correctness of LLMs' judgments on verifiably solvable instances. (1) A \textit{Correct Response ($+1$)} is assigned when the model outputs the correct answer for a solvable instance. (2) For \textit{Incorrect Reasoning ($0$)} on solvable instances, such as producing an incorrect final answer despite performing the calculation, the model receives a neutral reward ($0$). This ensures that honest calculation errors are not penalized, maintaining the model's incentive to attempt reasoning.\vspace{-5pt}

\paragraph{Reward for Unsolvability Detection ($R_{\text{detect}}$).} 
This component evaluates how accurately the model identifies inherently contradictory or ill-posed instances. 
(1) A \textit{Correct-Detect Response ($+1$)} is assigned when the model explicitly outputs the canonical tag \texttt{unsolvable} for such an instance. 
(2) To prevent degenerate behavior, we impose a \textit{False Unsolvability Penalty ($\rho = -0.5$)}: if the model incorrectly declares a solvable problem \texttt{unsolvable}, it receives a negative reward. This penalty effectively expands the decision threshold, encouraging the model to attempt reasoning even under moderate uncertainty. As derived in our decision-theoretic analysis in Appendix~\ref{app:proof1}, this term is mathematically critical to prevent the model from collapsing into a local optimum of \textit{universal rejection}. Moreover, our empirical ablation on $\rho$ selection identifies $\rho = -0.5$ as the optimal value; while stricter penalties yield marginal gains in solvable accuracy, they significantly impair unsolvable detection capabilities, making this value the most effective trade-off. We refer to this penalty mechanism as $P$ in our subsequent experimental analysis and tables.
(3) For \textit{Error-Detect Reasoning ($0$)} on unsolvable instances, such as hallucinating a solution instead of outputting \texttt{unsolvable}, the model receives a neutral reward.\vspace{-5pt}

\paragraph{Reward for Difficulty Calibration ($R_{\text{cal}}$).}
When the model outputs the specific refusal token (e.g., \texttt{beyond my capabilities}), we regulate refusal via a \textit{target accuracy threshold} $\tau \in [0,1]$. During each RL rollout, we compute the empirical accuracy $\beta$ of the current group:
\begin{equation}
    \beta = \frac{1}{N}\sum_{i=1}^{N} \mathbf{1}[y_i \text{ is correct}].
\end{equation}
The calibration reward is then defined as:
\begin{equation}
 R_{\text{cal}}(y) = \lambda (\tau - \beta)\,\mathbf{1}[y \text{ is a refusal}].
\end{equation}
The term $(\tau - \beta)$ quantifies the gap between desired and actual reliability. This yields a dynamic incentive: when the model's reasoning capability is insufficient ($\beta < \tau$), the reward becomes positive, \textbf{promoting refusal}; conversely, when performance is strong ($\beta > \tau$), it turns negative, thus \textbf{discouraging refusal} on tasks within its capability.

To enforce robust calibration, we implement a \textit{progressive schedule} for $\tau$, increasing it monotonically throughout training to approach $1$. This curriculum balances exploration (low $\tau$) and prudent strictness (high $\tau$). We provide a formal proof of this \textit{Capability-Refusal Trade-off} in Appendix~\ref{app:calibration_proof}, mathematically demonstrating why a static threshold leads to reward collapse and failure to refuse.

The final per-trajectory reward is the sum of these components: 
\begin{equation}
    R = R_{\text{acc}} + R_{\text{detect}} + R_{\text{cal}}.
\end{equation}
\begin{figure*}[t]
    \centering
    \includegraphics[width=1.0\linewidth]{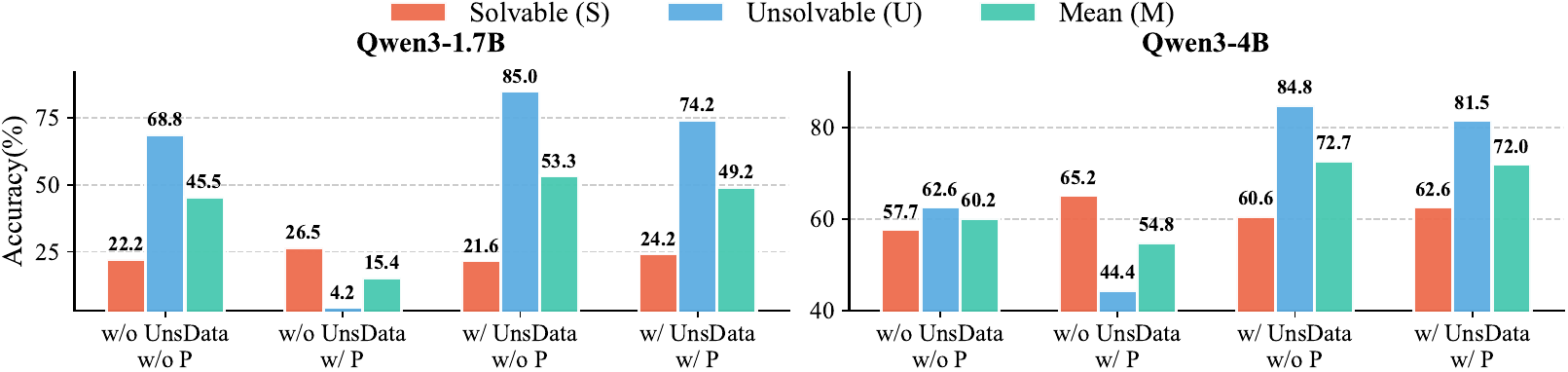}
    \caption{
        Detailed Ablation Metrics. 
        Breakdown of Solvable Accuracy (S, Red), Unsolvable Detection Rate (U, Blue), and Mean Performance (M, Green) across different configurations.
        The w/o UnsData (\textbf{Unsolvable Data}) w/ P setting shows a catastrophic drop in U for the 1.7B model (4.2\%), indicating capability collapse.
        Notably, w/ UnsData w/o P achieves the highest Mean (M) score by maximizing U.
        The full method (w/ UnsData w/ P) demonstrates a valuable trade-off, boosting S while maintaining a high U, optimizing for practical scenarios.
    }
    \label{fig:ablation_barplot}
\end{figure*}
\subsection{Answer Extraction and Verification}
Solvable tasks employ deterministic verifiers, including graph/constraint solvers and reference matchers. Unsolvable tasks are validated by the canonical \texttt{unsolvable} marker. Finally, responses containing ``beyond my capabilities'' trigger the dynamic calibration reward.

\section{Experiments}
\label{sec:results}


\begin{figure}[t]
    \centering
    \includegraphics[width=\linewidth]{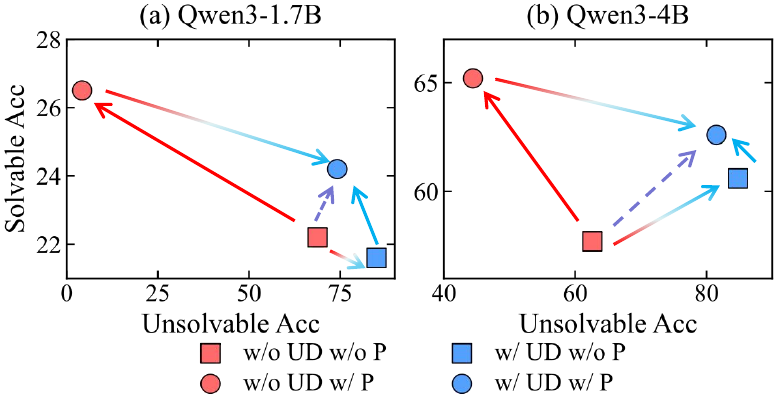}
    \caption{Visualizing the interaction between Solvable Accuracy ($S$) and Unsolvable Detection ($U$). 
    The Purple Arrow highlights our full method (Blue Circle) achieving a Pareto improvement over the baseline (Red Square). 
    Both models exhibit consistent dynamics: applying penalties without Unsolvable Data (UD) triggers \textbf{Capability Collapse} (Red Arrow), sacrificing $U$ to maximize $S$. 
    Conversely, when paired with UD, the penalty acts as a \textbf{Regularizer} (Blue Arrow), trading marginal $U$ for significant gains in $S$ compared to using UD alone (Blue Square).
    }
    \label{fig:ablation_scatter}
\end{figure}

We trained and evaluated Qwen3-1.7B and Qwen3-4B \citep{yang2025qwen3technicalreport} in think mode on UnsolvableQA. For evaluation, we report average@8 on the AIME partitions and average@4 on other datasets. We standardize training steps: ablation variants are trained for 120 steps, while the final models are trained for 320 steps.

\subsection{Main Results: Comparative Analysis with SOTA Models}
Table~\ref{tab:vertical_main_results_compact} compares our Qwen3-UnsolvableRL models against both proprietary SOTA models and base baselines. The complete experimental results are presented in Appendix~\ref{sec:appendix_comprehensive}, which also verifies the generalizability of our method on Llama models, where UnsolvableRL boosts the overall Mean score to 55.1\% compared to the 25.2\% baseline.
High-reasoning LLMs (e.g., Deepseek-V3.2-R1) inherently perform well, whereas base models struggle significantly (e.g., Qwen3-1.7B Instruct achieves only 32.4\% mean score). 
Our method bridges this gap: Qwen3-4B + UnsolvableRL achieves a global mean score of 78.6\%, significantly outperforming GPT-5.1-low (56.2\%) and approaching Gemini-3 (90.8\%). 
Notably, our 4B model achieves the highest global Unsolvable Detection rate (87.5\%), surpassing even the much larger Deepseek and Gemini models.
This confirms that equipping models with the ability to identify unsolvability ($\mathcal{U}$) strengthens the rigor required to solve feasible problems ($\mathcal{S}$). We further validated our method on two OOD benchmarks in Appendix~\ref{sec:appendix_ood}.
\begin{figure*}[t]
    \centering
    \includegraphics[width=1.0\linewidth]{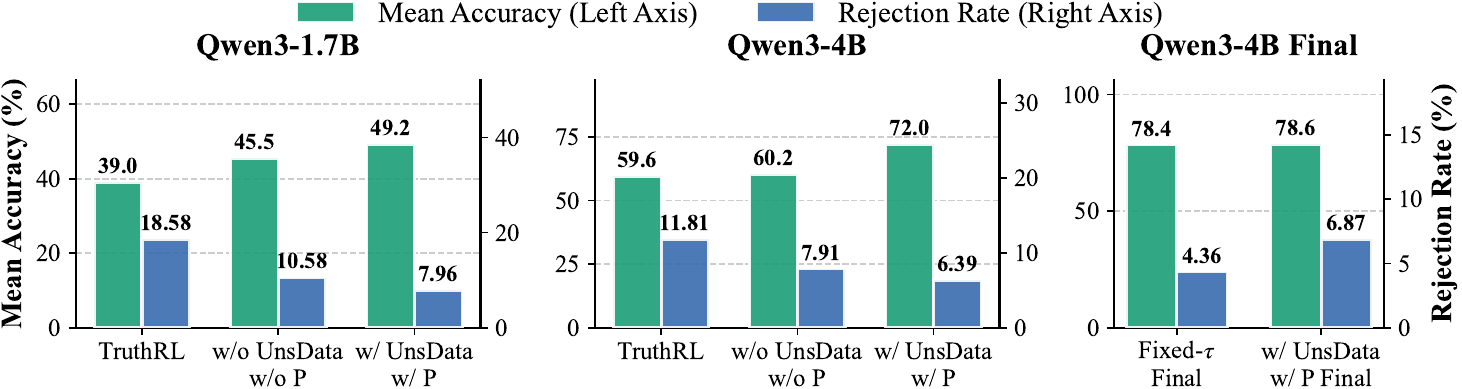}
    \caption{
        Performance Comparison and Calibration Analysis. 
        (Left/Middle) Comparison with TruthRL. While TruthRL achieves higher rejection rates (Blue bars), it suffers from over-refusal, significantly degrading Accuracy (Green bars).
        (Right) Comparison of threshold strategies. The adaptive mechanism (\textit{w/ P Final}) maintains a healthier rejection rate (6.87\%) compared to the Fixed-$\tau$ baseline (4.36\%) while preserving high accuracy.
    }
    \label{fig:performance_comparison}
\end{figure*}

\subsection{Ablation Study: Disentangling Data and Penalty}
To understand the contributions of data and reward mechanisms, we conducted a comprehensive ablation study comparing four configurations. The detailed metrics are shown in Figure~\ref{fig:ablation_barplot}, and the underlying dynamics are visualized in Figure~\ref{fig:ablation_scatter}.

\paragraph{Explicit Unsolvable Data is a Prerequisite for Detection.}
As shown in Figure~\ref{fig:ablation_barplot}, the naive baseline trained without unsolvable data (\textit{w/o UnsData}) lacks the fundamental capability to recognize unsolvability. For Qwen3-4B, detection is weak ($U=62.6\%$). Introducing explicit unsolvable data provides the necessary supervision, boosting detection significantly ($U \to 84.8\%$). This confirms that optimizing solely for reasoning correctness on solvable tasks is insufficient.

\paragraph{Applying Penalty without Data Triggers Capability Collapse.}
Figure~\ref{fig:ablation_scatter}(left) illustrates a critical failure mode caused by objective mismatch. When the penalty is enforced without providing ground-truth unsolvable examples, the model, lacking positive examples of valid refusal, minimizes the penalty risk by defaulting to answering all queries. This behavior effectively collapses its refusal mechanism, as seen in the 1.7B model's detection rate plummeting from 68.8\% to 4.2\%.

\paragraph{Applying Penalty with Data Facilitates a Strategic Trade-off.}
When grounded with adequate data, the penalty shifts from being a suppressor to a regularizer. For Qwen3-4B, adding the penalty (\textit{w/ UnsData w/ P}) improves Solvable Accuracy ($S$) from 60.6\% to 62.6\%, at the cost of a slight drop in detection ($U: 84.8\% \to 81.5\%$). This trade-off is strategic: it sacrifices marginal detection performance to enforce stricter reasoning rigor on feasible problems, optimizing the model for real-world utility.

\paragraph{Overall Gain: A Pareto Improvement over the Baseline.}
Viewing the trajectory globally (the purple dashed arrow in Figure~\ref{fig:ablation_scatter}b), our full method represents a comprehensive upgrade over the naive baseline (\textit{w/o UnsData w/o P}). We achieve significant gains in both dimensions for Qwen3-4B ($S: 57.7\% \to 62.6\%$, $U: 62.6\% \to 81.5\%$). This confirms that UnsolvableRL does not merely shift the decision boundary, but fundamentally enhances both reliability and reasoning capability.

\subsection{Comparative Analysis: TruthRL and Calibration}
We further benchmark our method against TruthRL~\citep{wei2025truthrl} and different thresholding strategies. 

\paragraph{Superior Balance Compared to TruthRL.}
As shown in the first two panels of Figure~\ref{fig:performance_comparison}, TruthRL tends to be overly conservative, leading to over-refusal. For Qwen3-1.7B, while TruthRL achieves a high Rejection Rate (18.58\%), its Accuracy drops to 39.0\%, which is inferior to even the baseline. In contrast, our method achieves a significantly higher Accuracy (45.5\%) with a moderate and precise Rejection Rate (10.58\%). This indicates that our \textit{target accuracy threshold} provides a softer, more distinctive signal than TruthRL's hard refusal reward, preserving the model's core reasoning capabilities.

\paragraph{Dynamic Thresholding Prevents Overconfidence.}
The rightmost panel of Figure~\ref{fig:performance_comparison} highlights the importance of our dynamic calibration. The Fixed-$\tau$ baseline yields a low rejection rate (4.36\%), suggesting overconfidence. By dynamically adjusting $\tau$ based on the moving average of solvable confidence, our method (w/ P Final) maintains a robust rejection rate (6.87\%) without sacrificing accuracy (78.6\%), ensuring the model remains prudent even as it becomes stronger.
\section{Conclusion}

In this paper, we address the challenge of LLM reliability by formally decoupling \textit{objective unsolvability} from \textit{subjective capability}. We introduce \textbf{UnsolvableQA} for rigorous evaluation and \textbf{UnsolvableRL}, an alignment framework featuring dynamic calibration. 
Empirically, our method achieves a \textbf{Pareto improvement} over baselines: it boosts solvable accuracy (Overall $S$) from 43.4\% to 69.4\% while maintaining robust unsolvability detection ($U > 85\%$ on Qwen3-4B). 
Crucially, our analysis uncovers the dual nature of alignment penalties: without grounding data, they trigger \textit{Capability Collapse}; yet with adequate data, they facilitate a \textit{Strategic Trade-off}, acting as a \textit{Regularizer for Rigor} that prevents hallucination without the over-refusal pitfalls observed in prior methods like TruthRL. 
We hope this work establishes a new paradigm for building models that rigorously distinguish between \textit{inherent unsolvability} and problems that simply \textit{exceed their current capabilities}.

\section*{Limitations}
While the UnsolvableQA dataset is relatively comprehensive within the selected domains (logic puzzles, mathematics, mazes), it may not fully encompass the various forms of inherently unsolvable problems encountered in real-world applications. Future work could expand the dataset to include more diverse and ambiguous types of unsolvable problems. Additionally, due to computational resource constraints, we were unable to conduct a more exhaustive set of ablation studies, such as scaling experiments on larger model architectures.

\bibliography{custom}

\appendix

\newpage
\appendix

\section{Appendix}
\subsection{Comprehensive experimental setup and results}

\begin{table*}[h]
\centering
\resizebox{\textwidth}{!}{
 \begin{tabular}{l *{24}{c}}
    \toprule
 & \multicolumn{3}{c}{Game24} & \multicolumn{3}{c}{HamCycle} & \multicolumn{3}{c}{HamPath} & \multicolumn{3}{c}{Hitori} & \multicolumn{3}{c}{Maze(E)} & \multicolumn{3}{c}{Maze(H)} & \multicolumn{3}{c}{AIME24-25} & \multicolumn{3}{c}{Overall} \\
    \textbf{Checkpoint} & S & U & M & S & U & M & S & U & M & S & U & M & S & U & M & S & U & M & S & U & M & $\mathrm{S}$ & $\mathrm{U}$ & $\mathrm{M}$ \\
\midrule
Deepseek-V3.2-R & 98.0 & 77.5 & 87.8 & 83.5 & 94.0 & 88.8 & 84.0 & 96.0 & 90.0 & 70.0 & 86.0 & 78.0 & 100.0 & 98.9 & 99.5 & 98.0 & 98.0 & 98.0 & 85.0 & 40.3 & 62.7 & 88.4 & 84.4 & 86.5 \\
GPT-5.1-Low & 14.0 & 50.0 & 32.0 & 8.3 & 82.0 & 45.1 & 50.0 & 86.0 & 68.0 & 6.0 & 12.0 & 9.0 & 95.0 & 100.0 & 97.5 & 86.0 & 94.0 & 90.0 & 61.7 & 42.5 & 52.1 & 45.9 & 66.6 & 56.2 \\
Gemini-3.0-Pro & 94.0 & 94.0 & 94.0 & 100.0 & 98.0 & 99.0 & 100.0 & 90.0 & 95.0 & 98.0 & 100.0 & 99.0 & 99.0 & 94.6 & 96.8 & 96.0 & 91.0 & 93.5 & 95.0 & 21.2 & 58.1 & 97.4 & 84.1 & 90.8 \\
\midrule
Qwen3-4B-Instruct & 81.0 & 17.0 & 49.0 & 37.5 & 57.0 & 47.3 & 37.0 & 56.5 & 46.8 & 34.5 & 6.5 & 20.5 & 32.7 & 55.6 & 44.1 & 13.3 & 64.5 & 38.9 & 67.9 & 14.8 & 41.4 & 43.4 & 38.8 & 41.1 \\
 + TruthRL & 91.0 & 91.0 & 91.0 & 32.3 & 86.0 & 59.2 & 49.0 & 90.0 & 69.5 & 38.0 & 15.0 & 26.5 & 90.0 & 77.1 & 83.6 & 31.0 & 56.5 & 43.8 & 67.5 & 19.1 & 43.3 & 57.0 & 62.1 & 59.6 \\
 + w/o UnsData w/o P & 83.0 & 92.0 & 87.5 & 36.5 & 81.0 & 58.8 & 47.0 & 84.0 & 65.5 & 46.0 & 20.0 & 33.0 & 88.5 & 76.6 & 82.6 & 34.5 & 60.5 & 47.5 & 68.3 & 24.0 & 46.2 & 57.7 & 62.6 & 60.2 \\
 + w/o UnsData w/ P & 90.5 & 12.5 & 51.5 & 39.6 & 67.5 & 53.5 & 49.5 & 64.5 & 57.0 & 56.5 & 4.0 & 30.3 & 94.8 & 83.2 & 89.0 & 56.3 & 72.0 & 64.1 & 69.0 & 7.4 & 38.2 & 65.2 & 44.4 & 54.8 \\
 + w/ UnsData w/o P & 84.0 & 95.0 & 89.5 & 39.6 & 96.0 & 67.8 & 53.0 & 100.0 & 76.5 & 45.0 & 87.0 & 66.0 & 95.5 & 95.2 & 95.4 & 40.0 & 89.5 & 64.8 & 67.1 & 30.8 & 49.0 & 60.6 & 84.8 & 72.7 \\
 + w/ UnsData w/ P & 88.0 & 97.0 & 92.5 & 35.9 & 95.5 & 65.7 & 51.5 & 99.5 & 75.5 & 48.5 & 71.0 & 59.8 & 94.0 & 95.5 & 94.8 & 53.3 & 79.8 & 66.6 & 66.9 & 31.9 & 49.4 & 62.6 & 81.5 & 72.0 \\
 + Fixed-$\tau$ Final & 87.5 & 99.0 & 93.3 & 39.1 & 100.0 & 69.6 & 67.0 & 99.5 & 83.3 & 51.0 & 98.0 & 74.5 & 96.3 & 100.0 & 98.2 & 52.5 & 95.8 & 74.2 & 69.0 & 41.9 & 55.5 & 66.1 & 90.6 & 78.4 \\
 + w/ UnsData w/ P Final & 94.5 & 99.0 & 96.8 & 41.1 & 94.5 & 67.8 & 55.0 & 96.5 & 75.8 & 63.5 & 94.5 & 79.0 & 96.5 & 98.9 & 97.7 & 65.8 & 89.0 & 77.9 & 69.6 & 40.4 & 55.0 & 69.4 & 87.5 & 78.6 \\
\midrule
Qwen3-1.7B-Instruct & 78.0 & 23.0 & 50.5 & 22.9 & 28.0 & 25.5 & 14.0 & 33.0 & 23.5 & 8.0 & 21.0 & 14.5 & 0.0 & 85.6 & 42.8 & 0.0 & 80.0 & 40.0 & 38.3 & 21.2 & 29.8 & 23.0 & 41.7 & 32.4 \\
 + TruthRL & 76.0 & 80.0 & 78.0 & 19.8 & 49.0 & 34.4 & 17.0 & 58.0 & 37.5 & 4.0 & 25.0 & 14.5 & 1.5 & 82.4 & 42.0 & 0.0 & 73.5 & 36.8 & 42.1 & 17.0 & 29.6 & 22.9 & 55.0 & 39.0 \\
 + w/o UnsData w/o P & 77.0 & 88.0 & 82.5 & 25.0 & 60.0 & 42.5 & 15.0 & 83.0 & 49.0 & 5.0 & 53.0 & 29.0 & 0.5 & 93.6 & 47.1 & 0.0 & 79.5 & 39.8 & 32.9 & 24.4 & 28.7 & 22.2 & 68.8 & 45.5 \\
 + w/o UnsData w/ P & 85.0 & 2.0 & 43.5 & 22.4 & 10.5 & 16.5 & 25.5 & 14.0 & 19.8 & 12.0 & 2.5 & 7.3 & 0.0 & 0.0 & 0.0 & 0.0 & 0.0 & 0.0 & 40.8 & 0.5 & 20.7 & 26.5 & 4.2 & 15.4 \\
 + w/ UnsData w/o P & 70.0 & 97.0 & 83.5 & 22.9 & 94.0 & 58.5 & 22.0 & 97.0 & 59.5 & 4.0 & 79.0 & 41.5 & 0.5 & 95.2 & 47.9 & 0.0 & 86.0 & 43.0 & 31.7 & 46.8 & 39.3 & 21.6 & 85.0 & 53.3 \\
 + w/ UnsData w/ P & 82.0 & 96.5 & 89.3 & 18.2 & 85.0 & 51.6 & 20.5 & 89.5 & 55.0 & 8.0 & 62.0 & 35.0 & 0.0 & 90.7 & 45.4 & 0.0 & 74.3 & 37.2 & 40.8 & 21.2 & 31.0 & 24.2 & 74.2 & 49.2 \\
 + w/ UnsData w/ P Final & 84.0 & 100.0 & 92.0 & 20.8 & 88.0 & 54.4 & 27.0 & 85.0 & 56.0 & 11.0 & 59.0 & 35.0 & 0.0 & 95.2 & 47.6 & 0.0 & 79.0 & 39.5 & 35.4 & 28.7 & 32.1 & 25.5 & 76.4 & 50.9 \\
\midrule
Llama3-8B-Distill & 28.0 & 48.0 & 38.0 & 27.1 & 2.0 & 14.6 & 14.0 & 0.0 & 7.0 & 6.0 & 56.0 & 31.0 & 1.0 & 44.7 & 22.9 & 0.0 & 52.0 & 26.0 & 35.0 & 39.4 & 37.2 & 15.9 & 34.6 & 25.2 \\
 + w/o UnsData w/o P  & 58.0 & 2.0 & 30.0 & 41.7 & 0.0 & 20.8 & 50.0 & 0.0 & 25.0 & 6.0 & 2.0 & 4.0 & 67.0 & 7.4 & 37.2 & 8.0 & 8.0 & 8.0 & 46.7 & 9.6 & 28.1 & 39.6 & 4.1 & 21.9 \\
 + w/ UnsData w/ P & 66.0 & 4.0 & 35.0 & 39.6 & 98.0 & 68.8 & 48.0 & 90.0 & 69.0 & 10.0 & 78.0 & 44.0 & 60.0 & 3.2 & 31.6 & 7.0 & 20.0 & 13.5 & 50.0 & 30.9 & 40.5 & 40.1 & 46.3 & 43.2 \\
 + w/ UnsData w/ P Final & 80.0 & 22.0 & 51.0 & 52.1 & 100.0 & 76.1 & 74.0 & 92.0 & 83.0 & 30.0 & 76.0 & 53.0 & 67.0 & 42.6 & 54.8 & 13.0 & 35.0 & 24.0 & 43.3 & 44.7 & 44.0 & 51.3 & 58.9 & 55.1 \\
\bottomrule
\end{tabular}}
\caption{Detailed performance breakdown (S/U/M) across all seven datasets. This table lists the full experimental results for all baselines and Qwen3 variants (including ablation checkpoints and final converged models). S: Solvable Accuracy; U: Unsolvable Detection Rate; M: Mean Score.}
\label{tab:results_per_domain_app}
\end{table*}

\begin{table*}[h]
\centering
\resizebox{\textwidth}{!}{
\begin{tabular}{l *{24}{c}}
\toprule
 & \multicolumn{3}{c}{Game24} & \multicolumn{3}{c}{HamCycle} & \multicolumn{3}{c}{HamPath} & \multicolumn{3}{c}{Hitori} & \multicolumn{3}{c}{Maze(E)} & \multicolumn{3}{c}{Maze(H)} & \multicolumn{3}{c}{AIME24-25} & \multicolumn{3}{c}{Overall} \\
        \textbf{Checkpoint} & Corr & Rej & C+R & Corr & Rej & C+R & Corr & Rej & C+R & Corr & Rej & C+R & Corr & Rej & C+R & Corr & Rej & C+R & Corr & Rej & C+R & Corr & Rej &C+R \\
\midrule
Deepseek-V3.2-R & 87.8 & 0.00 & 87.8 & 88.8 & 8.30 & 97.1 & 90.0 & 6.00 & 96.0 & 78.0 & 9.00 & 87.0 & 99.5 & 0.00 & 99.5 & 98.0 & 0.00 & 98.0 & 62.7 & 0.80 & 63.5 & 86.5 & 3.44 & 90.0 \\
GPT-5.1-Low & 32.0 & 1.00 & 33.0 & 45.1 & 2.08 & 47.2 & 68.0 & 8.00 & 76.0 & 9.0 & 1.00 & 10.0 & 97.5 & 0.00 & 97.5 & 90.0 & 0.00 & 90.0 & 52.1 & 7.90 & 60.0 & 56.2 & 2.85 & 59.1 \\
Gemini-3.0-Pro & 94.0 & 1.00 & 95.0 & 99.0 & 0.00 & 99.0 & 95.0 & 0.00 & 95.0 & 99.0 & 0.00 & 99.0 & 96.8 & 0.00 & 96.8 & 93.5 & 0.00 & 93.5 & 58.1 & 0.00 & 58.1 & 90.8 & 0.14 & 90.9 \\
\midrule
Qwen3-4B-Instruct & 49.0 & 0.00 & 49.0 & 47.3 & 24.74 & 72.0 & 46.8 & 12.00 & 58.8 & 20.5 & 4.00 & 24.5 & 44.1 & 0.00 & 44.1 & 38.9 & 0.25 & 39.2 & 41.4 & 0.00 & 41.4 & 41.1 & 5.86 & 47.0 \\
 + TruthRL & 91.0 & 0.50 & 91.5 & 59.2 & 34.15 & 93.4 & 69.5 & 16.00 & 85.5 & 26.5 & 18.50 & 45.0 & 83.6 & 2.50 & 86.1 & 43.8 & 11.00 & 54.8 & 43.3 & 0.00 & 43.3 & 59.6 & 11.81 & 71.4 \\
 + w/o UnsData w/o P & 87.5 & 0.00 & 87.5 & 58.8 & 28.85 & 87.7 & 65.5 & 15.50 & 81.0 & 33.0 & 11.00 & 44.0 & 82.6 & 0.00 & 82.6 & 47.5 & 0.00 & 47.5 & 46.2 & 0.00 & 46.2 & 60.2 & 7.91 & 68.1 \\
 + w/o UnsData w/ P & 51.5 & 0.00 & 51.5 & 53.5 & 21.09 & 74.6 & 57.0 & 7.00 & 64.0 & 30.3 & 2.25 & 32.6 & 89.0 & 0.00 & 89.0 & 64.1 & 0.13 & 64.2 & 38.2 & 0.00 & 38.2 & 54.8 & 4.35 & 59.2 \\
 + w/ UnsData w/o P & 89.5 & 1.50 & 91.0 & 67.8 & 19.00 & 86.8 & 76.5 & 10.00 & 86.5 & 66.0 & 13.00 & 79.0 & 95.4 & 1.00 & 96.4 & 64.8 & 4.00 & 68.8 & 49.0 & 0.42 & 49.4 & 72.7 & 7.00 & 79.7 \\
 + w/ UnsData w/ P & 92.5 & 1.00 & 93.5 & 65.7 & 19.75 & 85.5 & 75.5 & 15.50 & 91.0 & 59.8 & 7.50 & 67.3 & 94.8 & 0.00 & 94.8 & 66.6 & 1.00 & 67.6 & 49.4 & 0.00 & 49.4 & 72.0 & 6.39 & 78.4 \\
 + Fixed-$\tau$ Final & 93.3 & 0.00 & 93.3 & 69.6 & 18.49 & 88.1 & 83.3 & 12.00 & 95.3 & 74.5 & 0.00 & 74.5 & 98.2 & 0.00 & 98.2 & 74.2 & 0.00 & 74.2 & 55.5 & 0.00 & 55.5 & 78.4 & 4.36 & 82.8 \\
 + w/ UnsData w/ P Final & 96.8 & 0.00 & 96.8 & 67.8 & 26.82 & 94.6 & 75.8 & 19.00 & 94.8 & 79.0 & 2.25 & 81.3 & 97.7 & 0.00 & 97.7 & 77.1 & 0.00 & 77.1 & 55.0 & 0.00 & 55.0 & 78.6 & 6.87 & 85.5 \\
\midrule
Qwen3-1.7B-Instruct & 50.5 & 0.00 & 50.5 & 25.5 & 8.85 & 34.3 & 23.5 & 12.00 & 35.5 & 14.5 & 0.50 & 15.0 & 42.8 & 0.00 & 42.8 & 40.0 & 0.00 & 40.0 & 29.8 & 0.00 & 29.8 & 32.4 & 3.05 & 35.5 \\
 + TruthRL & 78.0 & 2.00 & 80.0 & 34.4 & 29.58 & 64.0 & 37.5 & 39.50 & 77.0 & 14.5 & 16.50 & 31.0 & 42.0 & 12.78 & 54.8 & 36.8 & 24.00 & 60.8 & 29.6 & 5.68 & 35.3 & 39.0 & 18.58 & 57.6 \\\
 + w/o UnsData w/o P & 82.5 & 0.50 & 83.0 & 42.5 & 29.17 & 71.7 & 49.0 & 30.00 & 79.0 & 29.0 & 6.50 & 35.5 & 47.1 & 0.25 & 47.3 & 39.8 & 3.50 & 43.3 & 28.7 & 4.17 & 32.9 & 45.5 & 10.58 & 56.1 \\
 + w/o UnsData w/ P & 43.5 & 0.25 & 43.8 & 16.5 & 19.01 & 35.5 & 19.8 & 8.00 & 27.8 & 7.3 & 5.25 & 12.5 & 0.0 & 0.00 & 0.0 & 0.0 & 0.00 & 0.0 & 20.7 & 0.00 & 20.7 & 15.4 & 4.64 & 20.0 \\
 + w/ UnsData w/o P & 83.5 & 1.00 & 84.5 & 58.5 & 22.88 & 81.4 & 59.5 & 19.50 & 79.0 & 41.5 & 2.50 & 44.0 & 47.9 & 1.50 & 49.4 & 43.0 & 0.75 & 43.8 & 39.3 & 2.52 & 41.8 & 53.3 & 7.24 & 60.5 \\
 + w/ UnsData w/ P & 89.3 & 0.00 & 89.3 & 51.6 & 27.77 & 79.4 & 55.0 & 19.50 & 74.5 & 35.0 & 2.50 & 37.5 & 45.4 & 0.00 & 45.4 & 37.2 & 4.50 & 41.7 & 31.0 & 1.46 & 32.5 & 49.2 & 7.96 & 57.2 \\
 + w/ UnsData w/ P Final & 92.0 & 0.50 & 92.5 & 54.4 & 32.19 & 86.6 & 56.0 & 21.50 & 77.5 & 35.0 & 2.00 & 37.0 & 47.6 & 0.55 & 48.2 & 39.5 & 6.75 & 46.3 & 32.1 & 3.17 & 35.3 & 50.9 & 9.52 & 60.4 \\
\midrule
Llama3-8B-Distill & 38.0 & 22.00 & 60.0 & 14.6 & 43.75 & 58.4 & 7.0 & 34.00 & 41.0 & 31.0 & 56.00 & 87.0 & 22.9 & 34.19 & 57.1 & 26.0 & 44.50 & 70.5 & 37.2 & 6.78 & 44.0 & 25.2 & 34.46 & 59.7 \\
 + w/o UnsData w/o P  & 30.0 & 0.00 & 30.0 & 20.8 & 5.21 & 26.0 & 25.0 & 4.00 & 29.0 & 4.0 & 6.00 & 10.0 & 37.2 & 1.00 & 38.2 & 8.0 & 4.00 & 12.0 & 28.1 & 0.83 & 28.9 & 21.9 & 3.01 & 24.9 \\
 + w/ UnsData w/ P & 35.0 & 0.00 & 35.0 & 68.8 & 16.67 & 85.5 & 69.0 & 3.00 & 72.0 & 44.0 & 19.00 & 63.0 & 31.6 & 3.00 & 34.6 & 13.5 & 8.50 & 22.0 & 40.5 & 0.83 & 41.3 & 43.2 & 7.29 & 50.5 \\
 + w/ UnsData w/ P Final & 51.0 & 2.00 & 53.0 & 76.1 & 10.42 & 86.5 & 83.0 & 4.00 & 87.0 & 53.0 & 19.00 & 72.0 & 54.8 & 4.05 & 58.9 & 24.0 & 17.00 & 41.0 & 44.0 & 3.34 & 47.3 & 55.1 & 8.54 & 63.7 \\
\bottomrule
\end{tabular}}
\caption{Alternative metric analysis showing Correctness (\textbf{Corr}), Rejection Rate (\textbf{Rej}), and their sum (\textbf{C+R}). This breakdown distinguishes between general accuracy and refusal tendency, providing a diagnostic view for over-refusal and capability collapse.}
\label{tab:agg_summary_horizontal_app_v2}
\end{table*}
\label{sec:appendix_comprehensive}

In this appendix, we provide the full, domain-level experimental results to supplement the aggregated metrics reported in the main text. Table~\ref{tab:results_per_domain_app} reports the performance breakdown across all seven datasets, including Game24, HamCycle, HamPath, Hitori, Maze-Easy, Maze-Hard, and AIME 24-25. We report three primary metrics to evaluate performance. First, Solvable Accuracy (S) measures the percentage of solvable problems correctly answered. Second, Unsolvable Detection Rate (U) indicates the percentage of unsolvable problems correctly identified as unsolvable. Finally, Mean (M) represents the arithmetic mean of S and U, reflecting the overall reliability of the model across different domains.

Complementing the standard metrics, Table~\ref{tab:agg_summary_horizontal_app_v2} provides a diagnostic view focusing on the model's refusal behavior. Correctness (Corr) measures the accuracy of the model's responses on attempted questions, independent of the refusal decision. Rejection Rate (Rej) tracks the proportion of total queries, including both solvable and unsolvable instances, that the model refused to answer. The sum of these two metrics (C+R) serves as a diagnostic tool to analyze trade-offs between reasoning capability and refusal tendency, helping to identify issues such as over-refusal or capability collapse.

Both tables include results for a wide range of configurations. We compare against strong proprietary models like Deepseek-V3.2-R, GPT-5.1-Low, and Gemini-3.0-Pro, as well as the base Qwen3-Instruct models, which serve as reference points. To isolate the effects of our proposed components, we evaluate ablation variants trained for 120 steps. These include models trained exclusively on solvable data without exposure to ground-truth unsolvable examples (denoted as w/o UnsData) and models trained without the False Unsolvability Penalty (denoted as w/o P), representing standard RLHF without the penalty term for incorrect refusals. Finally, we present results for the fully converged models trained for 320 steps, including a variant using a static confidence threshold (Fixed-tau Final) and our fully converged model trained with the complete UnsolvableRL method (w/ UnsData w/ P Final).

We also extended our evaluation to the Llama3-8B architecture to verify the generalizability of our method. As shown in the tables, training exclusively on solvable data (\textit{w/o UnsData w/o P}) yields a dramatic improvement in solvable accuracy compared to the baseline ($S: 15.9\% \to 39.6\%$). However, this comes at a catastrophic cost: the model's ability to detect unsolvable problems collapses ($U: 34.6\% \to 4.1\%$), leading to a net decline in the overall Mean score ($M: 25.2\% \to 21.9\%$). 

In contrast, our proposed method (\textit{w/ UnsData w/ P}) successfully resolves this conflict. At step 120, it maintains superior solvable accuracy ($S=40.1\%$) while simultaneously robustifying unsolvable detection ($U=46.3\%$), nearly doubling the overall Mean score to 43.2\%. \textbf{Furthermore, extending the training to approximately 400 steps (\textit{Final}) demonstrates robust scalability: the model achieves further gains with Solvable Accuracy reaching 51.3\% and Unsolvable Detection rising to 58.9\%, resulting in a comprehensive Mean score of 55.1\%.} This confirms that UnsolvableRL effectively prevents capability collapse and enhances reasoning rigor across different model families.

\subsection{Out-of-Distribution (OOD) Generalization}
\label{sec:appendix_ood}

To evaluate the generalization capabilities of UnsolRL, we utilized two Out-of-Distribution (OOD) benchmarks: \textbf{8-Puzzle} ($N_{sol}=87, N_{unsol}=100$) and \textbf{Zebra Logic} ($N_{sol}=100, N_{unsol}=100$). These tasks differ significantly from the training distribution, serving as a rigorous test for reasoning robustness. Table~\ref{tab:ood_single_column_trimmed} details the performance mean@4 across Solvable (\textbf{S}), Unsolvable (\textbf{U}), and Mean (\textbf{M}) metrics. 

\paragraph{SOTA Model Performance.}
We first established a baseline using top-tier proprietary models. \textbf{Deepseek-V3.2-R} achieves the highest overall performance (\textbf{M}=76.2\%), demonstrating exceptional dominance in logic tasks with a mean score of 92.7\% on Zebra Logic. In contrast, \textbf{Gemini-3.0-Pro} excels in procedural planning, securing the top spot on the 8-Puzzle benchmark (\textbf{M}=70.3\%). \textbf{GPT-5.1-Low} shows balanced capabilities, particularly in detecting unsolvable instances across both domains (Overall \textbf{U}=94.5\%). These results highlight that while leading models perform well, handling both logical constraints and path-finding robustly remains a non-trivial challenge.

\paragraph{Qwen3-4B Results.} 
The baseline Qwen3-4B model struggles significantly with OOD detection, failing completely to identify unsolvable 8-Puzzle instances (0.0\%). \textbf{UnsolRL-Final} substantially enhances robustness: it raises the \textbf{U} score to 15.8\% on the 8-Puzzle and achieves a remarkable improvement on Zebra Logic, boosting \textbf{U} from 69.0\% to 87.0\%. Consequently, the Overall Mean (\textbf{M}) increases by over 50\% relative to the baseline (from 23.2\% to 34.9\%), demonstrating that our method successfully transfers verification capabilities to unseen tasks.

\paragraph{Qwen3-1.7B Results.} 
Despite its limited capacity, the 1.7B model exhibits a similar positive trend in anomaly detection. On Zebra Logic, UnsolRL improves the unsolvable detection rate (\textbf{U}) from 36.0\% to 47.3\%. Although the complex 8-Puzzle remains highly challenging for this smaller architecture (Overall \textbf{M} for 8-Puzzle improves marginally to 1.1\%), the improvement in the Overall Mean across tasks (from 11.5\% to 13.8\%) confirms that UnsolRL effectively mitigates hallucinations on unsolvable queries, even when reasoning resources are constrained.

\begin{table*}[h]
\centering
\small 
\begin{tabular}{l *{3}{c} *{3}{c} *{3}{c}}
    \toprule
    & \multicolumn{3}{c}{\textbf{8-Puzzle}} & \multicolumn{3}{c}{\textbf{Zebra Logic}} & \multicolumn{3}{c}{\textbf{Overall}} \\
    \cmidrule(lr){2-4} \cmidrule(lr){5-7} \cmidrule(lr){8-10}
    \textbf{Model} & S & U & M & S & U & M & S & U & M \\
    \midrule
    Deepseek-V3.2-R  & 40.2 & 79.0 & 59.6 & \textbf{94.5} & \textbf{91.0} & \textbf{92.7} & \textbf{67.3} & 85.0 & \textbf{76.2}\\
    Gemini-3.0-Pro  & \textbf{43.6}& \textbf{97.0} & \textbf{70.3} & 65.0 & 74.0 & 69.5 &54.3 & 85.5 & 69.9 \\
    GPT-5.1-Low  & 28.9& 98.0  &  63.5 & 48.0  &  \textbf{91.0}&  69.5& 38.5 & \textbf{94.5} & 71.5 \\

    \midrule
    Qwen3-4B-Instruct & 21.6 & 0.0 & 10.8 & 2.0 & 69.0 & 35.5 & 11.8 & 34.5 & 23.2 \\
    + UnsolRL-Final   & \textbf{26.4} & \textbf{15.8} & \textbf{20.1} & \textbf{8.8} & \textbf{87.0} & \textbf{47.9} & \textbf{17.4} & \textbf{52.4} & \textbf{34.9} \\
    \midrule
    Qwen3-1.7B-Instruct & 0.8 & 0.0 & 0.4 & \textbf{9.0} & 36.0 & 22.5 & \textbf{4.9} & 18.0 &  11.5 \\
    + UnsolRL-Final     & \textbf{1.2} & \textbf{1.0} & \textbf{1.1} & 5.3 & \textbf{47.3} & \textbf{26.3} & 3.3 & \textbf{24.2} & \textbf{13.8} \\
    \bottomrule
\end{tabular}
\caption{OOD performance. \textbf{S}: Solvable, \textbf{U}: Unsolvable, \textbf{M}: Mean accuracy (\%).}
\label{tab:ood_single_column_trimmed}
\end{table*}
\subsection{Case Study}

To qualitatively demonstrate the effectiveness of our alignment strategy, we present three representative cases generated by \textbf{DeepSeek-V3.2-R}. These examples illustrate the model's ability to navigate complex reasoning paths: detecting latent contradictions in mathematics (Case 1), identifying topological impossibility in logic puzzles (Case 2), and maintaining flexible reasoning for solvable problems (Case 3).

\subsubsection*{Case 1: Latent Unsolvability (Number Theory)}

\paragraph{Problem Input.} 
Let $N$ be the greatest four-digit positive integer with the property that whenever one of its digits is changed to $1$, the resulting number is divisible by $7$. Suppose that $N \equiv Q+R \pmod{7}$, where $Q$ and $R$ are the quotient and remainder when $N$ is divided by $1000$. Find $Q+R$.

\paragraph{Original Model Response.}
\begin{quote}
\small
\textit{``The digit-change property requires that for any digit of $N$, replacing that digit with 1 results in a number divisible by 7. Under the strict interpretation... the only solutions are $N = 5624$ and $N = 5694$, with the latter being the greatest.
However, for $N = 5694$, computing $Q = 5$ and $R = 694$ gives $Q + R = 699$. Checking the congruence $N \equiv Q + R \pmod{7}$ yields $5694 \equiv 3 \pmod{7}$ and $699 \equiv 6 \pmod{7}$, which are not equal. ... Thus, the greatest $N$ satisfying the digit-change property does not satisfy the additional condition. Therefore, no four-digit $N$ meets all the requirements.''}
\[ \boxed{\text{unsolvable}} \]
\end{quote}

\paragraph{Reasoning Analysis.} 
This case exemplifies \textbf{"Verify-then-Conclude"}. The contradiction is not explicit in the problem statement but lies deep within the solution space.
\begin{itemize}
    \item \textbf{Deep Execution:} The model does not reject the problem early. It successfully solves the core number theory challenge, identifying the unique candidate $N=5694$.
    \item \textbf{Constraint Verification:} Instead of hallucinating a match, it rigorously computes the modular arithmetic ($3 \neq 6$) and identifies the conflict between the derived candidate and the final constraint.
\end{itemize}

\subsubsection*{Case 2: Topological Impossibility (Hitori)}

\paragraph{Problem Input.} 
Solve the following $4 \times 4$ Hitori puzzle. Rules: No duplicates in rows/cols; Black cells cannot touch; White cells must be connected.
\begin{center}
\small
\setlength{\tabcolsep}{4pt}
\begin{tabular}{|c|c|c|c|}
\hline
4 & 2 & \textbf{3} & \textbf{3} \\ \hline
2 & 1 & 4 & \textbf{2} \\ \hline
4 & 3 & 3 & 2 \\ \hline
4 & 1 & 2 & \textbf{2} \\ \hline
\end{tabular}
\end{center}

\paragraph{Original Model Response.}
\begin{quote}
\small
\textit{``The puzzle is unsolvable due to the isolation of cell (0,3) in all valid configurations... Specifically, the triplet of 2s in column 3 forces the outer cells (1,3) and (3,3) to be black to avoid adjacent shading. This forces (0,3) to be white. However, row 0 has duplicate 3s at (0,2) and (0,3), forcing (0,2) to be black. This leaves the white cell (0,3) surrounded by black cells (0,2) and (1,3), disconnecting it from the rest of the board.''}
\[ \boxed{\text{unsolvable}} \]
\end{quote}

\paragraph{Reasoning Analysis.}
This demonstrates the model's grasp of \textbf{Global Constraints}.
\begin{itemize}
    \item \textbf{Logical Deduction:} The model correctly deduces that to resolve duplicates in Col 3 (values: 3,2,2,2) and Row 0 (values: 4,2,3,3), specific neighbors of the corner cell (0,3) must be shaded.
    \item \textbf{Connectivity Check:} It recognizes that satisfying local uniqueness constraints inevitably violates the global connectivity rule (isolating the white cell at 0,3). This shows the model isn't just counting numbers but understanding the topological structure of the puzzle.
\end{itemize}

\subsubsection*{Case 3: Flexible Reasoning (Game24)}

\paragraph{Problem Input.} 
Use the numbers $\{4, 4, 6, 12\}$ to compute 24.

\paragraph{Original Model Response.}
\begin{quote}
\small
\textit{``To find an arithmetic expression that equals 24... One straightforward approach is to obtain a 4 from the numbers 12, 4, and 4, and then multiply the result by 6.
1. Start with 12. Subtract 4: $12 - 4 = 8$.
2. Subtract the second 4: $8 - 4 = 4$.
3. Multiply the result by 6: $4 \times 6 = 24$.
This gives the expression: $(12 - 4 - 4) \times 6$.''}
\[ (12 - 4 - 4) * 6 \]
\end{quote}

\paragraph{Reasoning Analysis.}
\begin{itemize}
    \item \textbf{No False Refusal:} Despite the rigorous training on unsolvable data, the model correctly identifies a valid path for a solvable query.
    \item \textbf{Flexibility:} The model finds a subtraction-based solution different from the standard division-based answer ($4 \times (12 / (6 - 4))$), showing that the alignment process has not constrained its exploration capability.
\end{itemize}

\paragraph{Summary.}
Together, these cases validate that DeepSeek-R aligned with UnsolvableRL does not rely on simple heuristics to reject questions. It performs \textbf{comprehensive reasoning}—whether calculating modular congruences, checking graph connectivity, or exploring arithmetic combinations—before making a solvability determination.

\begin{table*}[h]
\centering
\resizebox{\textwidth}{!}{
 \begin{tabular}{l *{24}{c}}
    \toprule
 & \multicolumn{3}{c}{Game24} & \multicolumn{3}{c}{HamCycle} & \multicolumn{3}{c}{HamPath} & \multicolumn{3}{c}{Hitori} & \multicolumn{3}{c}{Maze(E)} & \multicolumn{3}{c}{Maze(H)} & \multicolumn{3}{c}{AIME24-25} & \multicolumn{3}{c}{Overall} \\
    \textbf{Checkpoint} & S & U & M & S & U & M & S & U & M & S & U & M & S & U & M & S & U & M & S & U & M & $\mathrm{S}$ & $\mathrm{U}$ & $\mathrm{M}$ \\
\midrule
4B UnsolvableRL (P=-0.5, S40) & 79.0 & 93.0 & 86.0 & 29.2 & 98.0 & 63.6 & 48.0 & 98.0 & 73.0 & 37.0 & 41.0 & 39.0 & 53.0 & 91.0 & 72.0 & 13.0 & 78.0 & 45.5 & 70.0 & 31.9 & 51.0 & 47.0 & 75.8 & 61.4 \\
4B UnsolvableRL (P=-1.0, S40) & 74.0 & 90.0 & 82.0 & 35.4 & 98.0 & 66.7 & 34.0 & 98.0 & 66.0 & 50.0 & 36.0 & 43.0 & 48.0 & 74.5 & 61.2 & 22.0 & 78.0 & 50.0 & 68.3 & 29.7 & 49.0 & 47.4 & 72.0 & 59.7 \\
4B UnsolvableRL (P=-2.0, S40) & 80.0 & 90.0 & 85.0 & 29.2 & 98.0 & 63.6 & 40.0 & 98.0 & 69.0 & 42.0 & 22.0 & 32.0 & 61.0 & 70.7 & 65.8 & 15.5 & 76.5 & 46.0 & 67.5 & 29.7 & 48.6 & 47.9 & 69.4 & 58.6 \\
\bottomrule
\end{tabular}}
\caption{Detailed performance breakdown (S/U/M) across all seven datasets for different Penalty values at 40 steps.}
\label{tab:results_per_domain_app_}
\end{table*}
\subsection{Theoretical Justification and Empirical Ablation for False Unsolvability Penalty}

\subsection{Theoretical Justification and Empirical Ablation for False Unsolvability Penalty}
\label{app:proof1}

We explicitly motivate the penalty value $\rho = -0.5$ via a decision-theoretic analysis. Consider a model with an initial weak reasoning capability $\epsilon$ on solvable tasks (e.g., $\epsilon \approx 0.1$) and a posterior belief $p$ that a given problem is unsolvable. The model decides between two actions based on their expected rewards:

\textbf{(1) Attempting to Solve:} The expected reward depends on the problem being solvable and the model answering correctly: $\mathbb{E}[\text{Attempt}] \approx (1-p) \cdot \epsilon$.

\textbf{(2) Declaring Unsolvable:} The expected reward is the weighted sum of a correct rejection and a false rejection penalty: $\mathbb{E}[\text{Reject}] = p \cdot 1 + (1-p) \cdot \rho$.

The model will only attempt to reason if $\mathbb{E}[\text{Attempt}] > \mathbb{E}[\text{Reject}]$. Solving this inequality yields a critical threshold for the belief $p$:
\[
p < \frac{\epsilon - \rho}{1 + \epsilon - \rho}
\]
If we set $\rho=0$ (no penalty), the threshold collapses to $p < 0.09$, meaning the model refuses to answer unless it is over 91\% sure the problem is solvable. This leads to a local optimum of \textit{universal rejection}. By setting $\rho = -0.5$, we relax this threshold to $p < 0.375$. This ``Threshold Expansion'' effectively encourages the model to attempt reasoning even under moderate uncertainty, preventing the collapse of reasoning capabilities.

To empirically validate this theoretical derivation, we conducted an ablation study on the penalty magnitude using the Qwen3-4B model at 40 training steps, as detailed in Table \ref{tab:results_per_domain_app_}. We compared the default $\rho = -0.5$ against stricter penalties ($\rho = -1.0$ and $\rho = -2.0$). The results indicate that while the method is generally robust, continuing to decrease $\rho$ (making the penalty more severe) is counterproductive. Specifically, decreasing $\rho$ to $-2.0$ causes the Unsolvable Detection Rate (Overall U) to drop significantly from 75.8\% to 69.4\%, as the model becomes overly hesitant to declare unsolvability. Crucially, this sacrifice in detection yields only a marginal gain in Solvable Accuracy (Overall S increases from 47.0\% to 47.9\%), resulting in a decline in the global Mean score from 61.4\% to 58.6\%. Consequently, $\rho = -0.5$ proves to be the optimal value, striking the best balance between maintaining reasoning rigor and preserving the courage to reject unsolvable queries. To ensure reliability, we report the average results across two independent training runs, evaluated using mean@2.

\begin{algorithm}[t]
\caption{Game24 Generation and Verification}
\label{alg:game24}
\begin{algorithmic}[1]
\REQUIRE Difficulty level $D$ (determines count $k \in \{4,5,6\}$), Target $T=24$
\ENSURE A paired instance $(S, \text{Label})$
\STATE \textbf{Function} \textsc{Verify}($S$):
\STATE \quad $\mathcal{T} \leftarrow$ All binary expression trees for set $S$
\STATE \quad \textbf{for} each tree $t \in \mathcal{T}$ \textbf{do}
\STATE \quad \quad Evaluate $val \leftarrow t$ using \textbf{Rational Arithmetic}
\STATE \quad \quad \textbf{if} $val == T$ \textbf{then} \textbf{return} \textsc{Solvable}, $t$
\STATE \quad \textbf{end for}
\STATE \quad \textbf{return} \textsc{Unsolvable}, $\emptyset$
\STATE \textbf{End Function}
\REPEAT
\STATE Sample a set of $k$ integers $S \leftarrow \{n_1, \dots, n_k\}$ where $n_i \in [1, 13]$
\STATE $Status, Proof \leftarrow \textsc{Verify}(S)$
\STATE \textbf{if} goal is \textit{Solvable} \textbf{and} $Status == \textsc{Solvable}$:
\STATE \quad \textbf{return} $(S, Proof)$
\STATE \textbf{else if} goal is \textit{Unsolvable} \textbf{and} $Status == \textsc{Unsolvable}$:
\STATE \quad \textbf{return} $(S, \text{``Unsolvable''})$
\UNTIL{Valid instance found}
\end{algorithmic}
\end{algorithm}

\begin{table*}[h]
\centering
\resizebox{\textwidth}{!}{
\begin{tabular}{l *{24}{c}}
\toprule
 & \multicolumn{3}{c}{Game24} & \multicolumn{3}{c}{HamCycle} & \multicolumn{3}{c}{HamPath} & \multicolumn{3}{c}{Hitori} & \multicolumn{3}{c}{Maze(E)} & \multicolumn{3}{c}{Maze(H)} & \multicolumn{3}{c}{AIME24-25} & \multicolumn{3}{c}{Overall} \\
        \textbf{Checkpoint} & Corr & Rej & C+R & Corr & Rej & C+R & Corr & Rej & C+R & Corr & Rej & C+R & Corr & Rej & C+R & Corr & Rej & C+R & Corr & Rej & C+R & Corr & Rej &C+R \\
\midrule
4B UnsolvableRL ($\tau$=0.2, S40) & 90.5 & 3.00 & 93.5 & 63.7 & 23.21 & 86.9 & 72.5 & 10.50 & 83.0 & 32.5 & 6.50 & 39.0 & 69.0 & 8.75 & 77.7 & 46.5 & 21.00 & 67.5 & 49.5 & 0.00 & 49.5 & 60.6 & 10.42 & 71.0 \\
4B UnsolvableRL ($\tau$=0.4, S40) & 86.0 & 1.50 & 87.5 & 63.6 & 24.48 & 88.1 & 73.0 & 5.50 & 78.5 & 39.0 & 10.00 & 49.0 & 72.0 & 4.50 & 76.5 & 45.5 & 18.50 & 64.0 & 51.0 & 0.21 & 51.2 & 61.4 & 9.17 & 70.6 \\
4B UnsolvableRL ($\tau$=0.6, S40) & 86.0 & 3.00 & 89.0 & 58.2 & 31.19 & 89.4 & 66.0 & 13.50 & 79.5 & 39.5 & 8.00 & 47.5 & 71.3 & 6.75 & 78.1 & 47.8 & 26.50 & 74.3 & 48.0 & 0.83 & 48.8 & 59.5 & 12.82 & 72.3 \\
4B UnsolvableRL ($\tau$=0.8, S40) & 78.0 & 3.00 & 81.0 & 67.3 & 30.69 & 98.0 & 70.5 & 13.50 & 84.0 & 36.0 & 8.00 & 44.0 & 70.7 & 6.75 & 77.5 & 49.3 & 26.25 & 75.6 & 46.4 & 0.84 & 47.2 & 59.7 & 12.72 & 72.4 \\
4B UnsolvableRL ($\tau$=1.0, S40) & 85.5 & 3.50 & 89.0 & 59.7 & 35.19 & 94.9 & 65.5 & 21.50 & 87.0 & 26.5 & 24.50 & 51.0 & 78.8 & 2.75 & 81.6 & 47.0 & 26.25 & 73.3 & 49.3 & 0.42 & 49.7 & 58.9 & 16.30 & 75.2 \\
\bottomrule
\end{tabular}}
\caption{Detailed performance breakdown (Correctness/Rejection Rate/C+R) across all seven datasets for different $\tau$ values at 40 steps.}
\label{tab:agg_summary_horizontal_app_v3}
\end{table*}

\subsection{Theoretical Justification and Empirical Ablation for Capability-Refusal Dynamics}
\label{app:calibration_proof}

In this section, we derive the mathematical relationship between the model's reasoning capability and its tendency to refuse, explaining why a dynamic threshold is essential.

The model should choose to refuse only if the reward for refusal exceeds the expected reward of answering:
\begin{equation}
    R_{\text{cal}} > \mathbb{E}[R_{\text{ans}}].
    \label{eq:refusal_condition}
\end{equation}
Note that the expected reward for answering is always non-negative ($\mathbb{E}[R_{\text{ans}}] \ge 0$). 

\paragraph{Failure of Fixed Threshold.} If we were to use a \textit{fixed threshold} (e.g., $\tau=0.5$), as the model's reasoning capability improves during training, the batch accuracy $\beta$ will likely surpass $\tau$ (i.e., $\beta > \tau$). Once this occurs, the refusal reward becomes \textbf{negative} ($R_{\text{cal}} = \lambda(\tau - \beta) < 0$). 

Substituting this into Inequality~\ref{eq:refusal_condition}, the condition becomes $ \text{Negative Value} > \mathbb{E}[R_{\text{ans}}]$. Since $\mathbb{E}[R_{\text{ans}}] \ge 0$, this inequality is impossible to satisfy. Consequently, for well-performed batches, the model is incentivized to suppress refusal entirely to avoid penalties, effectively forcing the model to hallucinate answers on difficult problems rather than refusing. 

Our progressive schedule prevents this collapse by ensuring $\tau$ grows dynamically to maintain $\tau \gtrsim \beta$, keeping the refusal incentive positive for uncertain queries. Specifically, in our experiments, the Fixed-$\tau$ baseline uses a constant $\tau = 0.4$, while our dynamic schedule linearly increases $\tau$ from $0.4$ to $1.0$ over 400 training steps.

To further validate our threshold selection, we conducted a simple ablation study on $\tau$. As shown in Table~\ref{tab:agg_summary_horizontal_app_v3}, we found that when training for only 40 steps, setting $\tau=0.4$ generally yields strong Accuracy, while further decreasing $\tau$ has minimal impact on performance. Based on these observations, we adopted $\tau=0.4$ as the initial value for our main experiments.

\subsection{Logic Puzzle Generation Algorithms}

\subsubsection{Game24 Generation}
The Game24 task involves combining a set of integers using arithmetic operations $(+, -, \times, \div)$ to equal 24. To ensure absolute correctness, particularly for unsolvable instances, we employ an exhaustive search over all possible binary expression trees. Crucially, we utilize exact rational arithmetic (e.g., Python's \texttt{Fraction}) rather than floating-point calculations to prevent precision errors that could misclassify instances (e.g., $23.99999 \neq 24$).

For unsolvable instances, we use a rejection sampling strategy: we randomly sample integer sets and retain only those for which the exhaustive solver proves the solution set is empty.

\subsubsection{Hamiltonian Cycle and Path Generation}

The Hamiltonian Cycle problem asks whether a graph contains a closed loop visiting every vertex exactly once, while the Hamiltonian Path problem asks for a linear path visiting every vertex. Since these problems are NP-complete, we utilize a SAT-based verification pipeline to guarantee ground-truth labels.

\paragraph{SAT Encoding.} We encode the existence of a path/cycle as a Boolean Satisfiability Problem. Let $N$ be the number of vertices. We introduce boolean variables $x_{v,i}$ indicating that vertex $v$ occupies the $i$-th position in the sequence ($0 \le i < N$). The constraints are defined as follows: \textbf{Bijection} requires that $\sum_v x_{v,i} = 1$ for all $i$ and $\sum_i x_{v,i} = 1$ for all $v$. \textbf{Adjacency} mandates that for all $i \in [0, N-2]$, if $x_{u,i}$ and $x_{v,i+1}$ are true, then $(u,v)$ must be an edge in $E$. Finally, \textbf{Cycle Closure (Cycle Only)} ensures that if $x_{u, N-1}$ and $x_{v, 0}$ are true, then $(u,v) \in E$. We use the Minisat solver (via \texttt{pysat}) to determine satisfiability.

\paragraph{Unsolvability Injection.} Instead of purely random sampling, we inject structural properties that inherently prevent Hamiltonian traversals. These include: \textbf{Structural Bottlenecks}, where graphs are generated with ``bridge'' nodes or articulation points that partition the graph into components that cannot be traversed in a single pass; \textbf{Degree Constraints}, where we inject nodes with degree $<2$ (dead ends) for cycles, or ensure the presence of $>2$ nodes with degree 1 for paths; and \textbf{Disconnectivity}, where we explicitly construct graphs with multiple connected components.

\begin{algorithm}[t]
\caption{Hamiltonian Graph Generation}
\label{alg:hamiltonian}
\begin{algorithmic}[1]
\REQUIRE Type $T \in \{\text{Cycle}, \text{Path}\}$, Node count $N$
\ENSURE A graph $G=(V,E)$ with label
\STATE \textbf{Function} \textsc{VerifySAT}($G, T$):
\STATE \quad Construct CNF formula $\phi$ based on adjacency matrix of $G$
\STATE \quad Add bijection constraints to $\phi$
\STATE \quad \textbf{if} $T == \text{Cycle}$: Add closure constraint $v_{N-1} \leftrightarrow v_0$
\STATE \quad \textbf{return} \textsc{SAT-Solver}($\phi$)
\STATE \textbf{End Function}
\REPEAT
\STATE Initialize empty graph $G$ on $N$ nodes
\STATE \textbf{if} goal is \textit{Solvable}:
\STATE \quad Create a random permutation $P$ of vertices
\STATE \quad Add edges $(P_i, P_{i+1})$ to form a base Path/Cycle
\STATE \quad Add random noise edges to increase density
\STATE \textbf{else if} goal is \textit{Unsolvable}:
\STATE \quad Select strategy $\sigma \in \{\text{Disconnect}, \text{Bottleneck}, \text{DeadEnd}\}$
\STATE \quad Construct $G$ satisfying $\sigma$ (e.g., partition $V$ into $V_1, V_2$ with no edges between them)
\STATE \quad Add random noise edges strictly preserving $\sigma$
\STATE $IsFeasible \leftarrow \textsc{VerifySAT}(G, T)$
\UNTIL{($IsFeasible$ matches goal)}
\RETURN $G$
\end{algorithmic}
\end{algorithm}

\subsubsection{Hitori Generation}
Hitori is a logic puzzle played on an $N \times N$ grid. The goal is to shade cells such that no number appears more than once in any row or column, shaded cells do not touch orthogonally, and all unshaded cells remain connected.

We model the puzzle as a Constraint Satisfaction Problem (CSP). We define binary variables $x_{i,j} \in \{0, 1\}$ for each cell ($1$ denotes shaded). The key constraints are: \textbf{Uniqueness}, which requires that if a number $v$ appears $k$ times in any row/column, at least $k-1$ occurrences must be shaded; \textbf{Non-adjacency}, ensuring that for any adjacent cells $u, v$, $x_u + x_v \le 1$; and \textbf{Connectivity}, verified via BFS to ensure all cells where $x_{i,j}=0$ form a single connected component.

Solvable instances are filtered to ensure a \textit{unique} solution. Unsolvable instances are those where the CSP solver (combined with connectivity checks) returns a solution count of 0.

\begin{algorithm}[t]
\caption{Hitori Generation via CSP}
\label{alg:hitori}
\begin{algorithmic}[1]
\REQUIRE Grid size $N$
\ENSURE A validated Hitori grid $G$
\STATE \textbf{Function} \textsc{SolveHitori}($G$):
\STATE \quad Initialize CSP solver $P$
\STATE \quad Apply \textbf{Row/Col constraints}: Eliminate duplicates
\STATE \quad Apply \textbf{Adjacency constraints}: No adjacent black cells
\STATE \quad $\mathcal{S}_{local} \leftarrow P.\text{getSolutions()}$
\STATE \quad $\mathcal{S}_{valid} \leftarrow \emptyset$
\STATE \quad \textbf{for} each solution $s \in \mathcal{S}_{local}$ \textbf{do}
\STATE \quad \quad \textbf{if} \textsc{CheckConnectivity}($s$) \textbf{then} $\mathcal{S}_{valid}.\text{add}(s)$
\STATE \quad \textbf{end for}
\STATE \quad \textbf{return} $|\mathcal{S}_{valid}|$
\STATE \textbf{End Function}
\REPEAT
\STATE Generate random $N \times N$ grid $G$ with values $\in [1, N]$
\STATE $Count \leftarrow \textsc{SolveHitori}(G)$
\STATE \textbf{if} goal is \textit{Solvable} \textbf{and} $Count == 1$:
\STATE \quad \textbf{return} $G$ \COMMENT{Unique solution found}
\STATE \textbf{else if} goal is \textit{Unsolvable} \textbf{and} $Count == 0$:
\STATE \quad \textbf{return} $G$ \COMMENT{Provably no valid configuration}
\UNTIL{Valid instance found}
\end{algorithmic}
\end{algorithm}

\begin{algorithm}[t]
\caption{Maze Generation with Strategic Blockage}
\label{alg:maze}
\begin{algorithmic}[1]
\REQUIRE Dimensions $W, H$, Difficulty $D$
\ENSURE A maze grid $M$ and label $L$
\STATE \textbf{Function} \textsc{Verify}($M$):
\STATE \quad \textbf{return} \textsc{BFS}($M$, Start, End)
\STATE \textbf{End Function}
\REPEAT
\STATE $M_{base} \leftarrow$ \textsc{GenerateSolvableDFS}($W, H$)
\STATE \textbf{if} goal is \textit{Solvable}:
\STATE \quad \textbf{return} ($M_{base}$, \textsc{Verify}($M_{base}$))
\STATE \textbf{else if} goal is \textit{Unsolvable}:
\STATE \quad $\mathcal{P} \leftarrow$ All paths in $M_{base}$ from Start to End
\STATE \quad Select path $P \in \mathcal{P}$ based on difficulty $D$
\STATE \quad Select obstacle point $o \in P$ (excluding Start/End)
\STATE \quad $M_{blocked} \leftarrow M_{base} \cup \{o\}$
\STATE \quad \textbf{if} \textbf{not} \textsc{Verify}($M_{blocked}$):
\STATE \quad \quad \textbf{return} ($M_{blocked}$, \text{``Unsolvable''})
\UNTIL{Valid instance found}
\end{algorithmic}
\end{algorithm}

\begin{table}[t]
\centering
\caption{Hyperparameter settings used in our experiments. We employ Group Relative Policy Optimization (GRPO) with the KL divergence penalty explicitly disabled to encourage broader exploration within the solution space.}
\label{tab:hyperparameters}
\resizebox{0.48\textwidth}{!}{
\begin{tabular}{lc}

\toprule
\textbf{Hyperparameter} & \textbf{Value} \\
\midrule
\multicolumn{2}{c}{\textit{General Optimization \& Generation}} \\
\midrule
RL Algorithm & GRPO \\
Group Size ($G$) & 12 \\
KL Penalty Term & Disabled \\
PPO Mini-Batch Size & 8 \\
Sampling Temperature & 0.6 \\
Sampling Top-$k$ & 45 \\
Max Prompt Length & 2048 \\
Generation Engine  & VLLM \\
\midrule
\multicolumn{2}{c}{\textit{Infrastructure}} \\
\midrule
Hardware & 8 $\times$ NVIDIA A100 GPUs \\
Tensor Parallelism (TP) & 2 \\
\midrule
\multicolumn{2}{c}{\textit{Model-Specific Settings}} \\
\midrule
\textbf{Qwen3-1.7B} & \\
\quad Training Batch Size & $32\times12$ \\
\quad Max Response Length & 32,000 \\
\midrule
\textbf{Qwen3-4B} & \\
\quad Training Batch Size & $14\times12$ \\
\quad Max Response Length & 32,000 \\
\bottomrule
\end{tabular}}
\end{table}

\subsubsection{Maze Navigation Generation}

The Maze Navigation task requires finding a valid path from a start position (S) to an end position (E) in a grid-based maze, where movement is restricted by walls.

\paragraph{Verification Logic.} We employ Breadth-First Search (BFS) to determine solvability. BFS guarantees finding the shortest path if one exists. A maze is \textbf{Solvable} if a path from S to E exists, and \textbf{Unsolvable} if the set of reachable cells from S does not contain E.

\paragraph{Construction Pipeline.}
We first generate a complex solvable maze using a randomized Depth-First Search (DFS) algorithm, which ensures a spanning tree structure. To increase complexity, we selectively remove walls to introduce cycles and multiple potential paths, while validating that the solution remains unique or manageable using BFS.

To construct \textbf{Unsolvable Instances}, we employ a \textit{Strategic Blockage} method rather than random wall placement. We analyze the solvable maze to identify all valid paths from S to E, and then select a \textit{Critical Junction}, a coordinate $(x,y)$ that lies on a valid path (or is shared by multiple paths). We place an obstacle (wall) at $(x,y)$ and rigorously verify unsolvability using BFS. If the maze remains solvable (e.g., via a detour), we discard the instance or iteratively block remaining paths until connectivity is severed. This approach ensures that unsolvable mazes are structurally similar to solvable ones, often requiring the model to explore deep into the maze before realizing the path is blocked, thereby penalizing premature "give up" or hallucinated paths.

\subsection{Hyperparameters and Training Details}
\label{app:hyperparameters}
\begin{figure*}[t]
    \centering
    \includegraphics[width=0.9\linewidth]{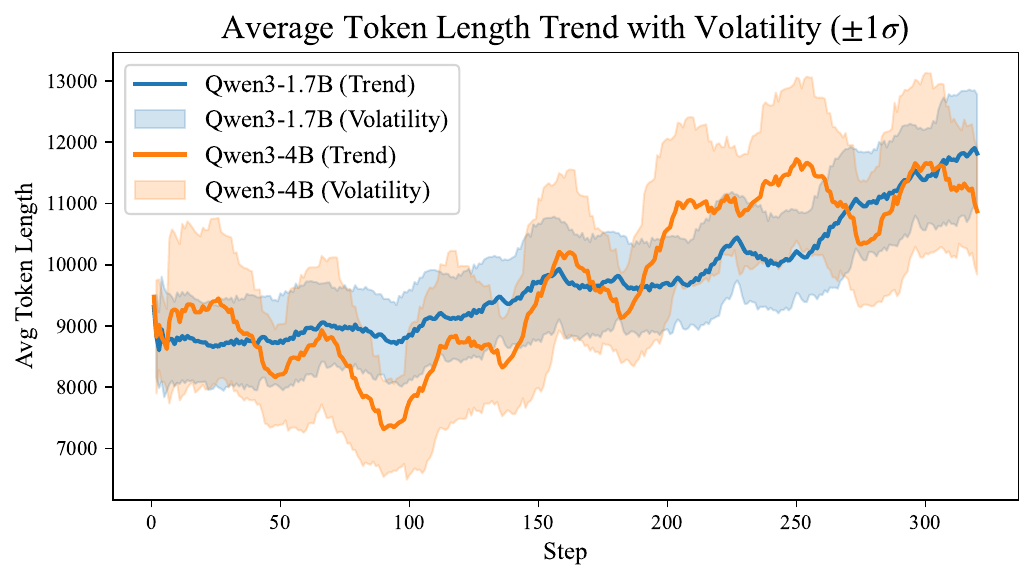}
    \caption{Average response token length trend during training. The solid lines represent the moving average, and the shaded regions indicate the volatility ($\pm 1\sigma$). Both models show a tendency to increase reasoning length as training progresses.}
    \label{fig:token_trend}
\end{figure*}
We implement our training pipeline using the \texttt{verl} (HybridFlow) framework and vllm engine~\citep{sheng2025hybridflow, kwon2023efficient}, which supports efficient reinforcement learning. The specific hyperparameters used for Group Relative Policy Optimization (GRPO) are detailed in Table~\ref{tab:hyperparameters}. We set the group size $G$ (denoted as \texttt{rollout.n} in \texttt{verl}) to 12. Notably, we explicitly set \texttt{use\_kl\_loss=False}, meaning the KL divergence penalty term was disabled during optimization, relying instead on the probability ratio clipping to maintain policy stability. 

We differentiate the batch size for Qwen3-1.7B and Qwen3-4B to accommodate their respective memory footprints. Specifically, the 1.7B model utilizes a larger batch size ($32 \times 12$) compared to the 4B model ($14 \times 12$). This adjustment balances the computational overhead, resulting in comparable training speeds for both architectures; typically, completing 120 training steps requires approximately one day on our 8 $\times$ NVIDIA A100 GPU cluster. Constrained by GPU resources, we conducted a single training run for the main experiments and report results using mean@8 for AIME and mean@4 for the remaining benchmarks.  Throughout the training process, we observed a stable increase in both Solvable Accuracy and Unsolvable Detection Rate.

To further analyze the training dynamics, Figure~\ref{fig:token_trend} illustrates the evolution of the average response token length throughout the training process. We observe a general upward trend for both models, indicating that the RL alignment encourages the generation of more comprehensive reasoning chains (longer CoT) to distinguish solvability. The Qwen3-4B model exhibits higher volatility (indicated by the wider shaded area) compared to the 1.7B model, suggesting more aggressive exploration within the solution space.

To explicitly guide the model's decision-making, we append specific instructions to the system prompt. 
For \textbf{unsolvable detection}, the instruction is: ``\textit{If you believe the problem is unsolvable, please output \texttt{\textbackslash boxed\{unsolvable\}} at the end.}'' 
For \textbf{capability calibration} (refusal), the instruction is: ``\textit{If your chain of thought is still confused after prolonged reasoning and you are likely to get the answer wrong, please provide the final output \texttt{\textbackslash boxed\{Beyond my capabilities\}}.}''. We rely on these predefined output formats to programmatically extract and classify the model's final response.

We will release our dataset, data construction pipeline, and training code upon acceptance.


\label{sec:appendix}

\end{document}